\pdfoutput=1

\documentclass[11pt]{article}

\usepackage{acl}

\usepackage{mathtools} 
\usepackage{makecell}
\usepackage{times}
\usepackage{latexsym}
\usepackage{amsmath}   
\usepackage{amsfonts}  
\usepackage{booktabs}  
\usepackage{listings}
\usepackage{xcolor}
\usepackage{mhchem} 
\DeclareUnicodeCharacter{2713}{\checkmark}

\usepackage[most]{tcolorbox}  
\usepackage{algorithm2e}
\usepackage{amssymb}     
\usepackage{verbatim}    
\usepackage{xcolor}      
\usepackage{hyperref}    

\usepackage{booktabs} 
\usepackage{colortbl}
\usepackage{multirow}
\usepackage{bbm}
\usepackage{enumitem}
\usepackage{makecell}

\usepackage[T1]{fontenc}

\usepackage[utf8]{inputenc}
\usepackage{amsmath}
\usepackage{amsthm}
\usepackage{enumitem}
\usepackage{multirow}
\usepackage{subcaption}
\usepackage{subfloat}
\usepackage{booktabs}

\usepackage{microtype}

\usepackage{inconsolata}

\usepackage{graphicx}

\usepackage{enumitem}
\usepackage{listings}  
\usepackage{pifont}
\usepackage[textsize=tiny]{todonotes}

\definecolor{darkgrey}{rgb}{0.53,0.53,0.53}
\definecolor{mygrey}{rgb}{0.9,0.9,0.9}
\definecolor{purple}{RGB}{230, 227, 254}
\definecolor{lightgreen}{RGB}{238, 252, 241}
\definecolor{lightred}{RGB}{231, 187, 187}
\definecolor{darkred}{RGB}{198, 129, 129}

\definecolor{tabhighlight}{HTML}{e5e5e5}
\definecolor{someorange}{rgb}{0.773,0.353,0.067}
\definecolor{someblue}{rgb}{0.27, 0.35, 0.760}

\usepackage{tabularx}
\usepackage{ragged2e}
\usepackage[most]{tcolorbox}

\definecolor{ExpertBG}{HTML}{EAF7EA}
\definecolor{StateBG}{HTML}{F5F5F7}
\definecolor{IWMBG}{HTML}{FDECF3}
\definecolor{SRBG}{HTML}{E6F2FF}
\definecolor{lightgray}{gray}{0.9}

\newtcolorbox{trainingexample}[2][]{%
  enhanced, breakable, colframe=black!12, colback=white, boxrule=2.5pt,
  arc=2pt, left=0pt, right=0pt, top=0pt, bottom=0pt,
  title={#2}, fonttitle=\bfseries, coltitle=black, #1}

\newcommand{\EXrow}[3]{\rowcolor{#1}\textbf{#2} & #3\\}

\usepackage{hyperref}

\usepackage{titlesec}   
\usepackage{titletoc}   
\usepackage{hyperref}
\usepackage[capitalize,noabbrev]{cleveref}

\usepackage{titletoc}
\definecolor{myred}{RGB}{0.8, 0, 0}

\titlecontents{lsection}
  [2em]                                    
  {\addvspace{8pt}\bfseries\color{myred}} 
  {\contentslabel{1.5em}}                  
  {}
  {\titlerule*[0.6pc]{.}\contentspage}    

\titlecontents{lsubsection}
  [4em]                                    
  {\color{myred}}                          
  {\contentslabel{2.3em}}                  
  {}
  {\titlerule*[0.6pc]{.}\contentspage}    

\usepackage{ifthen}
\usepackage{eso-pic}
\usepackage{graphicx}

\usepackage[capitalize,noabbrev]{cleveref}

\theoremstyle{plain}

\theoremstyle{definition}

\theoremstyle{remark}

\usepackage[textsize=tiny]{todonotes}

\title{TopoCurate: Modeling Interaction Topology for Tool-Use Agent Training}

\author{\
Jinluan Yang\textsuperscript{1}\thanks{These authors contributed equally. Corresponding authors: Qi Gu and Kun Kuang.}, Yuxin Liu\textsuperscript{2}\footnotemark[1], Zhengyu Chen\textsuperscript{2}\footnotemark[1], Chengcheng Han\textsuperscript{2}\footnotemark[1],\\
\textbf{Yueqing Sun\textsuperscript{2}, Qi Gu\textsuperscript{2}, Hui Su\textsuperscript{2}, Xunliang Cai\textsuperscript{2}, Fei Wu\textsuperscript{1}, Kun Kuang\textsuperscript{1}}\\
\textsuperscript{1}Zhejiang University \ \ \ \textsuperscript{2} Meituan LongCat\\
\texttt{yangjinluan@zju.edu.cn}\\
}

\begin{document}
\maketitle


\begin{abstract}
Training tool-use agents typically relies on outcome-based filtering: Supervised Fine-Tuning (SFT) on successful trajectories and Reinforcement Learning (RL) on pass-rate-selected tasks. However, this paradigm ignores interaction dynamics—successful trajectories may lack error recovery or exhibit redundancy, while pass rates fail to distinguish structurally informative tasks from trivial ones. To bridge this gap, we propose \textbf{TopoCurate}, an interaction-aware framework that shifts the paradigm from linear outcome filtering to topological interaction modeling. By projecting multi-trial rollouts into a unified semantic quotient topology within the same task, we construct a structured manifold that explicitly captures the divergence between success and failure pathways across tool invocations and environmental responses. 
Leveraging this representation, we introduce a dual-selection mechanism: for SFT, we prioritize trajectories demonstrating reflective recovery, semantic efficiency, and strategic diversity to mitigate covariate shift and mode collapse; for RL, we select tasks with high error branch ratios and strategic heterogeneity, maximizing gradient Signal-to-Noise Ratio to address vanishing signals in sparse-reward settings. Evaluations on BFCLv3 and Tau2 Bench show that TopoCurate achieves consistent gains of 4.2\% (SFT) and 6.9\% (RL) over state-of-the-art baselines.
\end{abstract}



\section{Introduction}

The advent of agentic intelligence has shifted the paradigm for large language models (LLMs) from static text generation to proactive engagement in multi-turn, tool-based interactions with environments\cite{team2025kimi,liu2025deepseek,yang2025qwen3,team2026longcat}. Consequently, the efficacy of agent training—whether via Supervised Fine-Tuning (SFT) or Reinforcement Learning (RL)—is strictly bottlenecked by the structural quality of the interaction data \cite{team2025mirothinker}. However, prevailing paradigms \cite{li2025simulating, xu2025toucan, yang2025toolmind} remain predominantly linear outcome-centric: SFT trajectories are selected based on binary success, and RL tasks by pass-rate thresholds from isolated trajectories. While scalable, these metrics are inherently myopic to interaction dynamics \cite{gooding2025interaction}, treating reasoning processes as flat, \textit{linear chains} and obscuring the critical \textit{structured dependencies}—such as error recovery and state transitions—that govern robust agency.

This reliance on outcomes creates a critical misalignment between data selection and learning value, a phenomenon we term the \textbf{Outcome Equivalence Illusion}. Our topological analysis reveals two core deficiencies:(i) \textit{SFT Selection Bias}: Standard outcome filtering selects successful trajectories indiscriminately, ignoring process robustness, efficiency, and strategic diversity. This dilutes critical error-recovery patterns with simple paths \cite{lan2025exploring}, while redundant loops and homogenized solutions lower training value \cite{zhu2025llm}. Consequently, the agent suffers from covariate shift, semantic inefficiency, and mode collapse \cite{yan2026pacevolve}, overfitting to narrow behaviors.(ii) \textit{RL Gradient Vanishing}: In sparse-reward RL, tasks with identical pass rates mask divergent structural training potentials. Tasks that yield homogeneous rollouts (identical outcomes regardless of action variations) produce vanishing policy gradients (low Advantage). In contrast, tasks with structural bifurcation—where sharp decision boundaries lead to outcome divergence—provide high-contrast signals, yielding a high Signal-to-Noise Ratio (SNR) for efficient optimization.

The root cause of this illusion is the prevailing \textit{linear, isolated} view of trajectories. Evaluating each rollout independently obscures the latent structure—the branching points, recovery paths, and state convergence—that defines robust agent behavior. To break this illusion, we propose \textbf{TopoCurate}, a framework that shifts the paradigm from linear outcome filtering to \textit{topological interaction modeling}. Our core innovation is to transcend the isolated analysis of trajectories by projecting multi-trial rollouts from the same task into a unified semantic topology. By merging semantically equivalent action-observation states across trials, we convert a sparse forest of linear traces into a structured, global manifold—a Directed Acyclic Graph (DAG). This topology explicitly visualizes the divergence between success and failure pathways and densifies visitation statistics, enabling the first rigorous, quantitative analysis of interaction process quality.


Leveraging this topological representation, we introduce a principled, dual-selection mechanism: For \textbf{SFT}, we define topological metrics for \textit{Reflective Recovery} (prioritizing error correction), \textit{Semantic Efficiency} (penalizing redundancy), and \textit{Distributional Diversity} (combating mode collapse) to curate a dataset that approximates a robust, strategic expert policy. For \textbf{RL}, we propose metrics for \textit{Error Branch Ratio} and \textit{Strategic Heterogeneity} to select tasks that maximize gradient SNR by prioritizing those with high structural complexity and diverse valid solutions, directly addressing the vanishing signal problem.


Our contributions are threefold:
\begin{itemize}
\item We introduce \textbf{TopoCurate}, a framework that shifts agent data curation from outcome-based filtering to topological interaction modeling. By projecting linear rollouts into a semantic quotient topology, we provide a rigorous formalism for capturing the interactions that outcome metrics neglect.

\item We devise a topology-driven dual-selection mechanism theoretically grounded in learning dynamics. For SFT, we introduce \textit{Reflective Recovery}, \textit{Semantic Efficiency}, and \textit{Distributional Diversity} to mitigate covariate shift and mode collapse. For RL, we propose structural task selection based on \textit{Error Branch Ratios} and \textit{Strategic Heterogeneity}, which maximize gradient SNR by prioritizing tasks with critical decision boundaries.
\item Extensive evaluations on BFCLv3 and Tau2 bench consistently demonstrate the effectiveness of our proposed method, achieving average gains of 4.2\% (SFT) and 6.9\% (RL) over state-of-the-art baselines.
\end{itemize}

\section{Related Work}

\subsection{Tool Use Data Synthesis and Selection}

Multi-Agent Simulation (MAS) is the prevailing paradigm for tool-use synthesis \cite{zhang2025nemotron}, encompassing approaches from social simulation \cite{tang2025synthesizing} to committee-based validation \cite{prabhakar2025apigen}. While systems like Kimi K2 \cite{team2025kimi} and Simia \cite{li2025simulating} scale MAS or simulate environmental feedback, their data selection remains heuristic and outcome-centric. Pipelines such as APIGen-MT \cite{prabhakar2025apigen} and Simia \cite{li2025simulating} rely on terminal checks or rule-based filtering, treating trajectories as flat sequences \cite{xu2025funreason}; similarly, MUA \cite{zhao2025mua} selects RL tasks solely by pass-rate. Although works like ToolPRM \cite{lin2025toolprm} and ToolPRMBench \cite{li2026toolprmbench} employ fine-grained process supervision, they focus on optimizing search or reward models rather than fundamentally assessing training data quality. Consequently, current paradigms neglect the interaction topology—its structural decision points and error recoveries—failing to separate robust process quality from stochastic success \cite{zhang2026robust}. In contrast, our TopoCurate framework addresses this gap by shifting to process-aware topological modeling, explicitly selecting data that maximizes structural learning value for both SFT and RL without external reward models.

\subsection{Tool-Use Evaluation}

The evaluation of tool-use agents has progressed from static syntax checks to dynamic, multi-turn interaction. Benchmarks like BFCL \cite{bfcl} and ACEBench \cite{chen2025acebench} assess multi-step tool invocation and error correction. Tau-Bench \cite{yao2024tau} and its successor $\tau_2$-bench \cite{barres2025tau} simulate dual-control environments where user and agent actions co-evolve the state, testing policy adherence in complex domains. Concurrently, evaluation metrics have shifted from greedy accuracy (Pass@1) to distributional robustness (Pass@k) \cite{yu2025pass, li2025simulating}. Pass@k measures an agent's capacity to discover diverse valid solutions within a stochastic search space, which serves in our work as a proxy for strategic heterogeneity—indicating whether the model has learned multiple pathways rather than overfitting to a single behavior.
\section{Preliminaries}
\label{sec:preliminaries}

In this section, we formalize the multi-turn tool-use process as a sequential decision-making problem and clarify the data selection of  SFT and RL training as our optimization object.

\subsection{Definition of Multi-Turn Agentic Tool Use}

\paragraph{Interaction Process.} 
We formulate the interaction between a tool-use agent and an environment as a Partially Observable Markov Decision Process \cite{kurniawati2022partially}. Define task as a tuple $\mathcal{T} = (q, \mathcal{C}, \mathcal{E})$, where $q$ denotes the user intent, $\mathcal{C}$ represents the context, and $\mathcal{E}$ is the evaluation protocol. At each step $t$, the agent receives an observation $o_t$ and generates an action unit $z_t = (r_t, a_t)$, comprising internal reasoning $r_t$ (e.g., Chain-of-Thought) and a structured tool action $a_t$. The interaction history is denoted as $h_t = (\mathcal{T}, z_1, o_1, \dots, z_{t-1}, o_{t-1})$. The goal is to learn a policy $\pi_\theta(z_t | h_t)$ that maximizes the task success.

\paragraph{Interaction Turns and Semantic Topology.} 
To model the agent-environment feedback loop, we define a complete interaction turn as $\hat{z}_t = (r_t, a_t, o_t)$, comprising the agent's reasoning, action, and the environmental observation. To capture the latent structure of tasks, we define a semantic equivalence relation over the space of complete turns $\hat{\mathcal{Z}}$. Two turns $\hat{z}_i, \hat{z}_j$ are considered equivalent if their actions and observations are semantically similar, indicating they represent the same logical state. This relation induces a quotient structure, compressing scattered linear trajectories into a unified Directed Acyclic Graph (DAG) where nodes represent clusters of semantically equivalent behaviors.

\subsection{Data Selection for Agentic Tool Use}

We consider a scenario with a large-scale task pool $\mathcal{Q}_{\text{pool}}$ and a set of synthesized trajectories. Our goal is to select high-quality subsets to optimize the agent's performance.

\paragraph{Trajectory Selection for SFT.}
Let $\mathcal{D}_{\text{pool}} = \{(\mathcal{T}_i, \tau_{i,j})\}$ be a dataset containing multiple candidate trajectories for each task. The data selection problem seeks a subset $\mathcal{D}_{\text{select}} \subset \mathcal{D}_{\text{pool}}$ (represented by weights $w_{i,j} \in [0, 1]$) to maximize validation performance under a budget constraint $B$:

\begin{equation}
\begin{aligned}
\max_{w}\quad &
    \mathbb{E}_{(\mathcal{T},\tau)\in\mathcal{D}_{\text{val}}}
    \bigl[R(\tau\mid \pi_{\theta^*})\bigr] \\
\text{s.t.}\quad &
    \theta^* = \arg\min_{\theta}
    \sum_{i,j} w_{i,j}\,
    \mathcal{L}_{\text{SFT}}(\pi_\theta,\tau_{i,j}), \\
&
    \sum_{i,j} w_{i,j}\le B.
\end{aligned}
\end{equation}


\noindent \textbf{Limitations.} Conventional filtering sets $w_{i,j} = \mathbb{I}[R(\tau_{i,j}) = 1]$. However, this linear outcome approach ignores the interaction topology that may lead to model collapse and covariate shift, e.g., A successful trajectory may still contain redundant loops, lack error recovery mechanisms, or represent a narrow solution path. The detailed discussions can be shown in Appendix \ref{app:theory_sft}. Our objective is to derive a topological prior $w_{i,j}$ that harmonizes execution optimality with process robustness, semantic efficiency, and strategic diversity.

\paragraph{Task Selection for RL.}
This objective is to curate a subset $\mathcal{Q}_{\text{RL}} \subset \mathcal{Q}_{\text{pool}}$ that maximizes gradient efficiency. Adopting Group Relative Policy Optimization (GRPO) as the training algorithm, the policy is updated by sampling a group of $K$ trajectories for each task and maximizing the relative advantage as Eq. \ref{eq:rl}, where $\hat{A}(\tau_k)$ is the standardized advantage from sparse outcome rewards.
\vspace{-2.5mm}


\begin{equation}
\begin{aligned}
J(\theta)
&= \mathbb{E}_{\mathcal{T}\sim P_{\mathrm{select}}(\mathcal{T})}
   \mathbb{E}_{\{\tau_{1:K}\}\sim\pi_\theta(\cdot\mid\mathcal{T})}\\
&\qquad \Biggl[\frac{1}{K}\sum_{k=1}^K
   \frac{\pi_\theta(\tau_k)}{\pi_{\mathrm{old}}(\tau_k)}
   \hat{A}(\tau_k)\Biggr]
\end{aligned}
\label{eq:rl}
\end{equation}

\noindent \textbf{Limitations.} Standard approaches define $P_{\text{select}}(\mathcal{T})$ based solely on pass rates. However, this overlooks structural training potential. Tasks yielding homogeneous rollouts produce vanishing advantages ($\hat{A} \approx 0$), providing minimal gradient information (low Signal-to-Noise Ratio, SNR). The detailed discussions can be shown in Appendix \ref{app:theory_rl}. Our framework constructs $P_{\text{select}}(\mathcal{T})$ to prioritize tasks with high Error Branch Ratios and Strategic Heterogeneity, maximizing the contrastive signal of the policy gradient within the topology.

\section{Methodology}
\label{sec:method}

\subsection{Overview}

\label{sec:overview}
\begin{figure*}[t]
    \centering
    \includegraphics[width=0.9\linewidth]{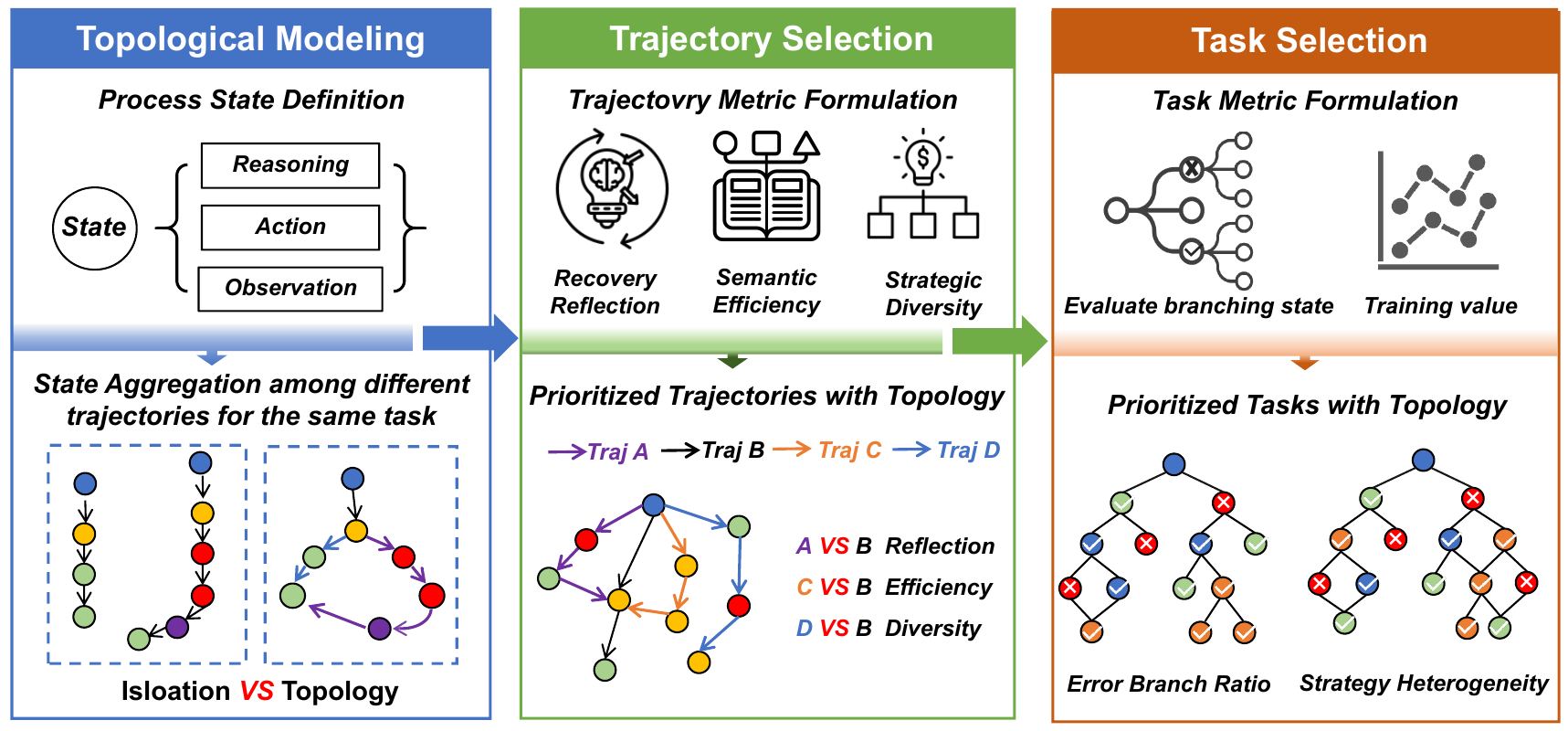}
    \caption{\textbf{Overview of the TopoCurate Framework.} Our method operates in three stages: \textbf{(Left) Topological Modeling} transforms disjoint rollouts into a unified state-transition graph by defining states via action-observation tuples and aggregating semantically equivalent turns; \textbf{(Middle) Trajectory Selection for SFT} applies three process-aware metrics—Reflective Recovery (resilience), Semantic Efficiency (economy), and Strategic Diversity (exploration)—to prioritize high-quality trajectories; \textbf{(Right) Task Selection for RL} evaluates tasks using Error Branch Ratio and Strategy Heterogeneity to select structurally complex tasks that maximize gradient efficiency. }
    \label{fig:method}
\end{figure*}

As illustrated in Figure \ref{fig:method}, our proposed framework operates in three systematic stages:
\textbf{(i) Topological Modeling (Sec. \ref{sec:topology}):} We aggregate raw interaction turns into a Directed Acyclic Graph (DAG), defining states by both tool actions and environmental observations. This structure reveals the \textit{Success Potential} of each state;
\textbf{(ii) Trajectory Selection for SFT (Sec. \ref{sec:sft}):} Instead of simple outcome filtering, we select trajectories based on three topological process metrics—\textit{Reflective Recovery} (resilience), \textit{Semantic Efficiency} (economy), and \textit{Strategic Diversity} (exploration)—to distill a robust expert policy;
\textbf{(iii) Task Selection for RL (Sec. \ref{sec:rl}):} We prioritize tasks that exhibit high structural complexity. By filtering for high \textit{Error Branch Ratios} and \textit{Strategic Heterogeneity}, we select tasks that provide sharp decision boundaries and diverse solution paths, maximizing training efficiency in sparse-reward settings.

\vspace{-2mm}
\subsection{Modeling Agentic Dynamics via Quotient Topology}
\label{sec:topology}

Unlike static reasoning chains, agentic tool use is driven by the environment's feedback loop. While the agent outputs $z_t=(r_t, a_t)$, the semantic state is determined jointly by the action and the environment's response $o_t$. To rigorously evaluate this process, we define a \textit{complete interaction turn} as $\hat{z}_t = (r_t, a_t, o_t)$ and compress the space of these turns $\hat{\mathcal{Z}}$ into a dense state space $\mathcal{V}$ via a quotient map $\pi: \hat{\mathcal{Z}} \to \mathcal{V}$.

\paragraph{Why Merge? The Necessity of State Aggregation.}
In long-horizon agentic tasks, the action space is combinatorial, and slight variations in tool arguments result in a disjoint, sparse forest. This sparsity prevents statistical analysis. By defining a semantic equivalence relation and merging equivalent turns, we achieve two critical goals:
\begin{itemize}
    \item \textit{Densification for Potential Estimation}: Aggregating sparse visits into dense node clusters allows us to robustly estimate the Success Potential Field $\Phi(v) = P(\text{Success}|v)$. Without merging, $\Phi(v)$ would be binary or undefined.
    \item \textit{Revealing Causal Convergence}: Merging reveals states reached by diverse paths. This ``equifinality" uncovers the underlying causal structure of the environment, enabling us to compare the efficiency of different strategies (e.g., shortcuts vs. detours) that arrive at the same intermediate state.
\end{itemize}

\paragraph{Observation-Driven State Abstraction.}
A key insight is that agent state is critically defined by observation $o_t$. For instance, in retail scenarios, \texttt{modify\_order(id="123")} leads to distinct states depending on whether $o_t$ returns \texttt{"Success"} or \texttt{"Error: User Confirmation Required"}. Let $\text{sim}(\cdot, \cdot)$ denote the cosine similarity between embeddings. We define the equivalence relation $\hat{z}_i \sim \hat{z}_j$ as a conjunction of view-specific constraints:
\begin{equation}
\begin{split}
    \hat{z}_i \sim \hat{z}_j \iff & \left( \text{sim}_{\text{tool}}(a_i, a_j) \ge \delta_{\text{tool}} \right) \\
    & \land \left( \text{sim}_{\text{result}}(o_i, o_j) \ge \delta_{\text{result}} \right).
\end{split}
\end{equation}

We utilize a high-performance embedding model (e.g., \texttt{jina-embeddings-v2-base-en \cite{gunther2023jina} and Qwen3-0.6B-Embedding} \cite{zhang2025qwen3}) with default thresholds ($\delta_{\text{tool}} \approx 0.95, \delta_{\text{result}} \approx 0.90$) to ensure the reasonability of quotient topology. More detailed robustness analysis can be shown in Appendix \ref{appdendix_embedding}.

\subsection{Topological Trajectory Selection for SFT}
\label{sec:sft}

Standard SFT filtering selects all trajectories with $R(\tau)=1$, implicitly assuming a uniform quality distribution. To solve the constrained optimization problem defined in Eq. (2), we propose a second-stage refinement that approximates the optimal weights $w(\tau)$ using topological signals. 

\paragraph{Metric Formulation.}
We derive three topological signals to quantify process quality:
\begin{itemize}
    \item \textit{Reflective Recovery ($S_{\text{ref}}$)}: Prioritizes trajectories where the agent recovers from negative feedback (Dip-Recovery Pattern). A reflection event occurs when potential drops ($\Phi(v_{t}) \ll \Phi(v_{t-1}) - \epsilon_{\text{dip}}$) and is subsequently restored:
    \begin{equation}
        S_{\text{ref}}(\tau) = \sum_{t \in \text{jumps}} \frac{\Phi(v_{t+k}) - \Phi(v_{t})}{k}.
    \end{equation}
This metric captures agent resilience in dual-control dynamics. For instance, in a telecom task, a trajectory encountering a Destination Unreachable error but recovering via device reboot yields high $S_{\text{ref}}$, whereas abandoning the task incurs penalty.
    
    \item \textit{Semantic Efficiency ($S_{\text{eff}}$)}: Penalizes redundant loops by comparing actual path length to the geodesic distance on the graph:
    \begin{equation}
        S_{\text{eff}}(\tau) = \min_{(u,v) \in \text{Convergence}(\tau)} \frac{\mathcal{D}_{\mathcal{G}}(u, v)}{\text{len}(\tau_{u \to v})}.
    \end{equation}
    This serves as a regularizer. In an airline booking context, it penalizes agents that redundantly query \texttt{get\_flight\_details} for the same ID without state changes, favoring concise information gathering.

    \item \textit{Distributional Diversity ($S_{\text{rare}}$)}: Maximizes entropy by weighting the inverse branch popularity $P(v)$ with the local success rate $\text{Acc}(v)$:
    \begin{equation}
        S_{\text{rare}}(\tau) = \frac{1}{|\tau|} \sum_{v \in \tau} \frac{\text{Acc}(v)}{\log(1 + P(v))}.
    \end{equation}
    This metric prevents mode collapse by prioritizing trajectories that traverse rare but successful decision branches, ensuring the policy learns diverse strategies.
\end{itemize}

\paragraph{Trajectory Selection Strategy.}
To balance these objectives, we define the selection weight $w(\tau)$ as:
\begin{equation}
    w(\tau) = \lambda_{\text{ref}} \cdot \tilde{S}_{\text{ref}}(\tau) + \lambda_{\text{rare}} \cdot \tilde{S}_{\text{rare}}(\tau) + \lambda_{\text{eff}} \cdot \tilde{S}_{\text{eff}}(\tau),
\end{equation}
where $\tilde{S}$ denotes the normalized score (via z-score standardization across the candidate pool), and $\{\lambda_{\text{ref}}, \lambda_{\text{rare}}, \lambda_{\text{eff}}\}$ are hyperparameters that control the relative importance of each metric. We select trajectories based on their composite scores $w(\tau)$, ensuring a balanced representation of robustness, diversity, and efficiency. Detailed hyperparameter sensitivity analysis are provided in Appendix \ref{app:hyperparams}.

\subsection{Topological Task Selection for RL}
\label{sec:rl}

Standard task selection based solely on pass rates overlooks structural complexity. We propose a topology-driven task selection mechanism.

\paragraph{Metric Formulation.}
We evaluate the structural complexity of each task's solution space:
\begin{itemize}
    \item \textit{Error Branch Ratio ($V_{\text{struct}}$)}: The proportion of decision branches leading to failure (where potential drops below a threshold $\epsilon_{\text{fail}}$), representing contrastive difficulty:
    \begin{equation}
    \small
        \resizebox{0.9\linewidth}{!}{$\displaystyle
        V_{\text{struct}}(\mathcal{T}) = \frac{1}{|\mathcal{B}|} \sum_{v \in \mathcal{B}} \frac{|\{u \in \text{children}(v) : \Phi(u) < \epsilon_{\text{fail}}\}|}{|\text{children}(v)|}.
        $}
    \end{equation}
    High $V_{\text{struct}}$ indicates critical decision nodes. In a retail ``Modify Order" task,  the state bifurcates at the \texttt{check\_status} node: respecting the ``Shipped" status leads to success, while ignoring it triggers an irreversible failure branch. Such sharp contrasts provide high-fidelity gradient signals for policy adherence.

    \item \textit{Strategic Heterogeneity ($V_{\text{div}}$)}: The Unique Chain Ratio (UCR), measuring the number of valid distinct tool sequences. High UCR indicates multiple valid workflows, preventing mode collapse:
    \begin{equation}
        V_{\text{div}}(\mathcal{T}) = \frac{|\{\tau : R(\tau) = 1\}|}{|\{\tau : \text{sampled}\}|}.
    \end{equation}
    For example, a flight rebooking request might be solvable via two distinct topological paths: a direct \texttt{modify\_flight} call or a composite \texttt{cancel} followed by \texttt{book}. Training on such tasks fosters policy plasticity and generalization.
\end{itemize}

\paragraph{Task Selection Strategy.}
We construct the task sampling distribution $P_{\text{select}}(\mathcal{T})$ to favor structural richness:
\begin{equation}
    P_{\text{select}}(\mathcal{T}) \propto \exp \left( \frac{V_{\text{struct}}(\mathcal{T}) + \alpha V_{\text{div}}(\mathcal{T})}{T} \right),
\end{equation}
where $\alpha$ is a balancing coefficient and $T$ is the temperature.

\section{Experiments}
\label{sec:experiments}

\subsection{Experimental Setup}
\label{sec:setup}

\paragraph{Backbones.}
We employ the Qwen3 series \cite{yang2025qwen3} (thinking mode, scaling from 8B to 32B) as our foundational backbones. 


\paragraph{Seed Data Construction.}
We construct the initial pool using a pipeline adapted from APIGen-MT \cite{prabhakar2025apigen}, tailored to Tau2 Environment. Tasks are stratified by Qwen3-32B's pass rate: fundamental tasks (high pass rate) are distilled into expert trajectories via Claude-4.5-Sonnet \cite{anthropic2024claude4} to seed the SFT pool, while complex tasks (moderate pass rate) are reserved for the RL pool. Crucially, this generation establishes a raw candidate set. TopoCurate is subsequently applied to select SFT trajectory and RL tasks. The Details are provided in Appendix \ref{appdendix_data}.

\begin{table}[t]
    \centering
    \caption{\textbf{Baseline Comparison.} All methods start with outcome-based filtering; TopoCurate further refines via topological metrics.}
    \label{tab:baseline_comparison}
    \setlength{\tabcolsep}{3.5pt}
    \resizebox{\linewidth}{!}{
    \begin{tabular}{l|c|c}
        \toprule
        \textbf{Method} & 
        \makecell{\textbf{Trajectory}\\\textbf{Selection (SFT)}} & 
        \makecell{\textbf{Task}\\\textbf{Selection (RL)}} \\
        \midrule
        APIGEN-MT \cite{prabhakar2025apigen} & 
        Outcome ($r=1$) & 
        N/A \\
        \midrule
        MUA \cite{zhao2025mua} & 
        Outcome ($r=1$) & 
        Pass Rate \\
        \midrule
        Simia-Tau \cite{li2025simulating} & 
        Outcome ($r=1$) & 
        N/A \\
        \midrule
        \rowcolor{gray!15}
        \textbf{TopoCurate (Ours)} & 
        \makecell{Outcome ($r=1$)\\$\downarrow$ \textbf{Topological}\\$S_{ref}, S_{eff}, S_{rare}$} & 
        \makecell{Pass Rate\\$\downarrow$ \textbf{Topological}\\$V_{struct}, V_{div}$} \\
        \bottomrule
    \end{tabular}
    }
    \vspace{-8pt}
\end{table}

\paragraph{Baselines.} We select baselines related to $\tau$-bench environment: 
\textbf{(i) APIGEN-MT} \cite{prabhakar2025apigen}, which employs LLM committee validation for successful trajectories; 
\textbf{(ii) MUA} \cite{zhao2025mua}, which selects trajectories via outcome filtering; 
and \textbf{(iii) Simia-Tau} \cite{li2025simulating}, which applies rule-based post-validation for correctness. While all methods perform outcome-based filtering as the primary selection criterion, TopoCurate adopt a \textbf{two-stage refinement} (Table \ref{tab:baseline_comparison}). After the initial outcome filter following previous works, we construct a quotient topology from the candidate set and apply process-aware metrics ($S_{ref}, S_{eff}, S_{rare}$ for SFT; $V_{struct}, V_{div}$ for RL) to select data with high structural quality. This design rigorously isolates the contribution of topological curation and verifies its flexibility.


\paragraph{Evaluation Benchmarks.}
We evaluate our approach on two challenging benchmarks: (i) \textbf{Tau2 Bench} (\textit{In-Domain}). For each test, we conduct 8 repeated tests to achieve pass@1, pass@4, pass@8 for stable evaluation; and (ii) \textbf{BFCL v3 Multi-Turn} (\textit{Out-of-Domain}), the Berkeley Function Calling Leaderboard, in order to keep the same settings with baselines \cite{zhao2025mua} to assess the model's generalization across unseen APIs and long interactions. Moreover, we report the results on the advanced \textbf{BFCL v4} in Appendix \ref{appendix_bfcl} to further verify the effectiveness of our method.

\subsection{Main Results}
\label{sec:main_results}


\begin{table*}[t]
    \centering
    \footnotesize  
\caption{Performance comparison on Tau2 Bench (IID) and BFCL v3 Multiturn (OOD). Models are trained on Tau2 environments. Gray rows denote TopoCurate variants. \textbf{Bold} indicates best results.}
    \begin{tabular}{l|ccc|cccc}
        \toprule
        \multirow{2}{*}{\textbf{Model}} & 
        \multicolumn{3}{c|}{\textbf{Tau2 Bench (IID)}} & 
        \multicolumn{4}{c}{\textbf{BFCL V3 Multiturn (OOD)}} \\
        \cmidrule(lr){2-4} \cmidrule(lr){5-8}
        & Airline & Retail & Telecom & Base & Long Context & Miss Func & Miss Param \\
        \midrule
        \multicolumn{8}{c}{\textbf{32B Size Models}} \\  
        \midrule
        \textbf{Qwen3-32B-Thinking} & 0.435 & 0.525 & 0.248 & 0.340 & 0.210 & 0.155 & 0.245 \\
        \midrule
        APIGen-MT-SFT & 0.347 & 0.592 & 0.202 & 0.375 & 0.215 & 0.175 & 0.205 \\
        MUA-SFT       & 0.311 & 0.582 & 0.193 & 0.350 & 0.195 & 0.210 & 0.285 \\
        MUA-RL        & 0.454 & 0.673 & 0.283 & 0.420 & 0.215 & 0.200 & 0.300 \\
        \rowcolor{gray!10}
        \midrule
        TopoCurate-SFT (w/o Topology) & 0.483 & 0.591 & 0.521 & 0.410 & 0.235 & 0.125 & 0.245 \\
        \rowcolor{gray!10}
        TopoCurate-SFT                & 0.526 & 0.623 & 0.539 & 0.430 & 0.265 & 0.135 & 0.295 \\
        \rowcolor{gray!10}
        TopoCurate-RL (w/o Topology)  & 0.531 & 0.643 & 0.721 & 0.450 & 0.245 & 0.165 & 0.260 \\
        \rowcolor{gray!10}
        TopoCurate-RL                 & \textbf{0.558} & \textbf{0.687} & \textbf{0.784} & \textbf{0.480} & \textbf{0.295} & \textbf{0.225} & \textbf{0.325} \\
        \midrule
        \multicolumn{8}{c}{\textbf{8B Size Models}} \\  
        \midrule
        \textbf{Qwen3-8B-Thinking} & 0.297 & 0.389 & 0.244 & 0.230 & 0.135 & 0.045 & 0.135 \\
        \midrule
        APIGen-MT-SFT & 0.229 & 0.491 & 0.126 & 0.245 & 0.140 & 0.050 & 0.170 \\
        MUA-SFT       & 0.160 & 0.314 & 0.090 & 0.240 & 0.100 & 0.110 & 0.165 \\
        MUA-RL        & 0.190 & 0.498 & 0.218 & 0.210 & 0.110 & 0.115 & 0.150 \\
        Simia-Tau-SFT & 0.395 & 0.522 & 0.145 & 0.045 & 0.020 & 0.005 & 0.080 \\
        Simia-Tau-RL  & 0.405 & 0.529 & 0.156 & 0.040 & 0.025 & 0.015 & 0.075 \\
        \rowcolor{gray!10}
        \midrule
        TopoCurate-SFT (w/o Topology) & 0.436 & 0.500 & 0.418 & 0.265 & 0.135 & 0.075 & 0.150 \\
        \rowcolor{gray!10}
        TopoCurate-SFT                & 0.445  & 0.529 & 0.435 & 0.265 & 0.125 & 0.090 & 0.160 \\
        \rowcolor{gray!10}
        TopoCurate-RL (w/o Topology)  & 0.467 & 0.518 & 0.533 & 0.255 & 0.145 & 0.105 & 0.160 \\
        \rowcolor{gray!10}
        TopoCurate-RL                 & \textbf{0.481} & \textbf{0.545} & \textbf{0.595} & \textbf{0.295} & \textbf{0.160} & \textbf{0.135} & \textbf{0.180} \\
        \bottomrule
    \end{tabular}
    \label{tab:main_results_clean}
    \vspace{-3mm}
\end{table*}

\paragraph{TopoCurate can consistently achieve better performance and generalization than previous methods due to the topological refinement.} As shown in Table \ref{tab:main_results_clean}, our empirical evaluation confirms that TopoCurate consistently establishes new state-of-the-art performance, outperforming leading baselines on both in-domain and OOD benchmarks. Specifically, (i) \textit{Superiority over In-Domain Baselines:} Despite sharing the same $\tau$-bench generation environment as APIGen-MT and Simia-Tau, TopoCurate-SFT achieves higher Pass@1 scores across all domains. This validates our core premise that \textit{data quality, defined by topological richness, outweighs mere generation quantity}. By filtering out redundant or spurious trajectories, TopoCurate distills a denser training signal, enabling robust learning from fewer examples; and (ii) \textit{Generalization to OOD Tasks:} On the BFCL v3 benchmark, our method demonstrates superior generalization (e.g., lower parameter miss rates), indicating that the reasoning primitives learned from topologically curated data are transferable to unseen APIs, rather than being mere domain heuristics.

\begin{figure*}[t]
    \centering
    \begin{subfigure}[b]{0.9\textwidth} 
        \centering
        \includegraphics[width=\linewidth]{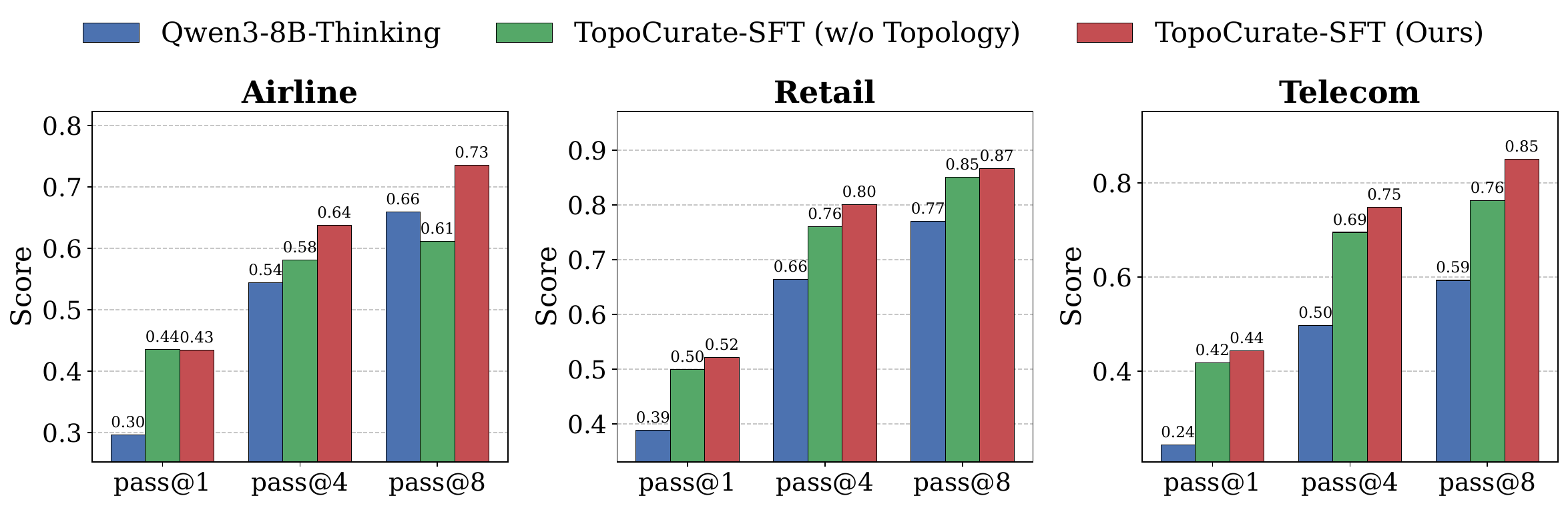}
        \vspace{-15pt}
        \label{fig:passk_8b}
    \end{subfigure}
    \vspace{-1pt}
    \begin{subfigure}[b]{0.9\textwidth}
        \centering
        \includegraphics[width=\linewidth]{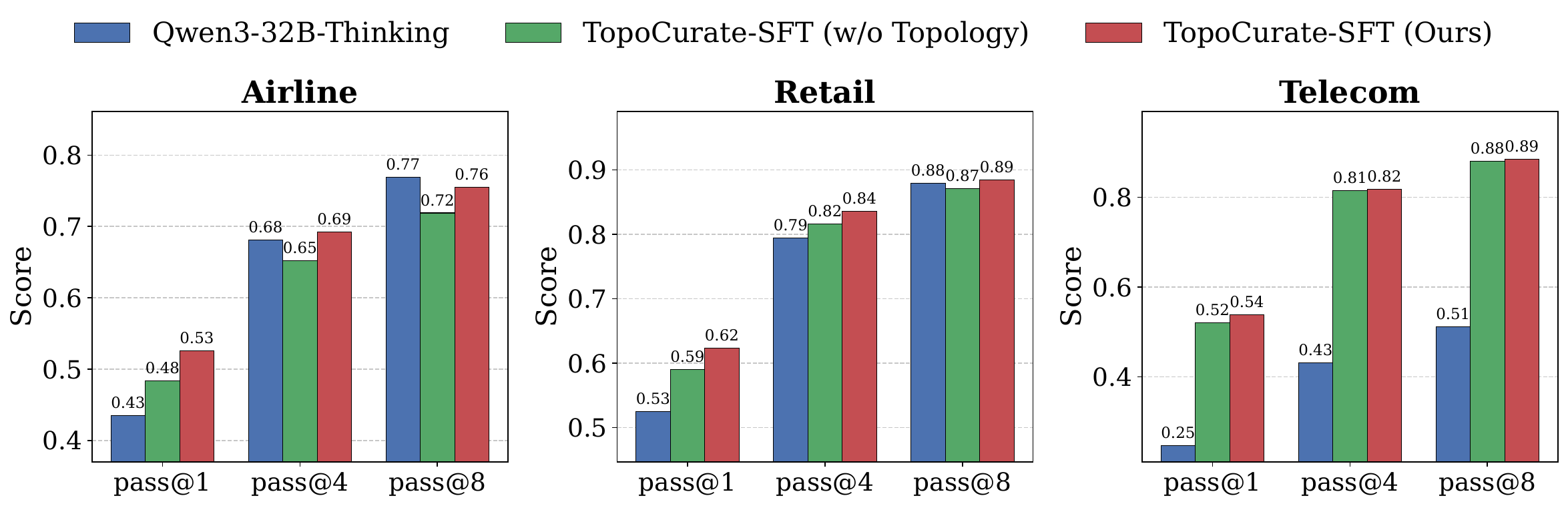}
        \vspace{-15pt}
        \label{fig:passk_32b}
    \end{subfigure}
    \caption{\textbf{Pass@k Comparison across three settings}: the Qwen3-Instruct backbone (Thinking mode), a baseline SFT model trained on standard outcome-filtered data (w/o Topology), and our TopoCurate-SFT. The results demonstrate that TopoCurate consistently achieve higher pass@k across all domains and model scales.}
    \label{fig:main_results}
\end{figure*}
\paragraph{Pass@k Analysis and  Scaling Behavior.}
We evaluate Pass@k performance across sampling budgets (Figure \ref{fig:main_results}) to validate TopoCurate's robustness. Key findings include:(i) Superior Intrinsic Capability. TopoCurate-SFT achieves substantially higher Pass@1 baselines across all domains. For 32B models, gains over the base model reach +23\% (Airline: 0.53 vs. 0.43), +17\% (Retail: 0.62 vs. 0.53), and +116\% (Telecom: 0.54 vs. 0.25). Similar improvements hold for 8B models (+43\% Airline, +33\% Retail, +83\% Telecom), demonstrating that topological curation fundamentally enhances model capability rather than merely improving sampling efficiency.(ii) Consistent Gains at Higher Budgets. As sampling increases (Pass@4, Pass@8), TopoCurate maintains its advantage. In Telecom (32B), TopoCurate's Pass@8 (0.89) outperforms w/o Topology (0.88) and substantially exceeds the base model (0.51), indicating enhanced distribution quality over viable solution paths. These results validate that topological curation distills robust expert policies capable of adaptive navigation across diverse solution trajectories. Further evidence from Appendix \ref{fig:rl_training_reward} corroborates these findings.

\subsection{Ablation Studies}
\label{sec:ablation}

\textbf{Trajectory Selection for SFT:} We dissect the contribution of our three topological SFT metrics (Table \ref{tab:ablation_components}). Removing \textit{Reflective Recovery} consistently degrades performance, validating that ``dip-recovery" patterns are essential for resilience. \textit{Efficiency} plays a critical role in Retail, where its removal drops accuracy by 6.1\% (0.529 $\to$ 0.468), while \textit{Diversity} is pivotal for Airline, with a 5.2\% drop (0.445 $\to$ 0.393). The full TopoCurate method harmonizes these objectives, achieving the highest average accuracy (0.470), confirming that an optimal policy must balance resilience, diversity, and economy; \textbf{Task Selection for RL:}
As shown in Table \ref{tab:ablation_rl}, we observe that filtering by \textit{Structural Complexity} ($V_{\text{struct}}$) is the dominant factor. Tasks with high error branch ratios provide the necessary contrastive signals—distinguishing optimal actions from critical failure modes—that are absent in trivial tasks. \textit{Strategic Heterogeneity} ($V_{\text{div}}$) acts as a vital regularizer, preventing the policy from overfitting to a narrow set of solution paths. 
\vspace{-3mm}

\subsection{Model Behavior and Training Dynamics}
\label{sec:behavior_dynamics}


\paragraph{Model Behavior Analysis.}
We analyze three behavioral dimensions as Figure \ref{fig:behavior_analysis} in Appendix. 
(i) \textit{Reflection:} TopoCurate-SFT agents exhibit markedly higher reflection scores, particularly in Telecom where the reflection rate increases from $\sim$0.15 to $\sim$0.28, confirming the internalization of self-correction mechanisms; (ii) \textit{Efficiency:} We observe a shift from fragile brevity to robust persistence. Measured by turns per success (lower is better), our method reduces interaction turns across domains by roughly \(\sim\)0.5–2 turns per success, indicating higher efficiency at a superior success rate; (iii) \textit{Strategic Plasticity:} The consistently higher Unique Chain Ratio indicates that our topological diversity metric effectively mitigates mode collapse, fostering a policy capable of adaptively selecting from a diverse repertoire of tool chains.

\paragraph{RL Training Dynamics Analysis.} 
We conduct two controlled experiments to validate topological curation's impact on  SFT initialization and  RL task selection from the training dynamic perspective. (i) \textit{SFT Initialization.}
We compare RL runs from different SFT checkpoints (TopoCurate-selected vs. outcome-filtered) using the same RL task pool. Figure \ref{fig:rl_training_reward} (left) shows the topologically-initialized model achieves a higher starting reward (0.29 vs. 0.22) and superior results (0.63 vs. 0.61 at step 60). Evaluation accuracy (Figure \ref{fig:rl_eval_accuracy}, left) confirms consistent gains across domains; (ii) \textit{ RL Task Selection.}
Starting from the same SFT checkpoint, we compare topologically-selected tasks versus uniform sampling. Figure \ref{fig:rl_training_reward} (right) reveals identical starting points ($\sim$0.25) but divergent outcomes (0.65 vs. 0.61 at step 60). Evaluation accuracy (Figure \ref{fig:rl_eval_accuracy}, right) shows w/ Topology consistently outperforms across domains (Airline: 0.545 vs. 0.535; Retail: 0.642 vs. 0.632; Telecom: 0.565 vs. 0.551 at step 60). These experiments across two stages consistently verify the effectiveness of TopoCurate.

\begin{table}[t]
    \centering
    \caption{\textbf{Ablation Study (SFT).} Impact of topological metrics on Pass@1 accuracy (Qwen3-8B-Thinking as backbone). ``Ref", ``Div", and ``Eff" denote Reflection, Diversity, and Efficiency.}

    \resizebox{\linewidth}{!}{
    \begin{tabular}{l|ccc|cccc}
        \toprule
        \multirow{2}{*}{\textbf{Method}} & \multicolumn{3}{c|}{\textbf{Components}} & \multicolumn{4}{c}{\textbf{Tau2 Pass@1}} \\
        & \textbf{Ref} & \textbf{Div} & \textbf{Eff} & \textbf{Airline} & \textbf{Retail} & \textbf{Telecom} & \textbf{Avg} \\
        \midrule
        TopoCurate-SFT (w/o Topology) & $\times$ & $\times$ & $\times$ & 0.436 & 0.500 & 0.418 & 0.451 \\ 
        \midrule
        w/o Reflection & $\times$ & $\checkmark$ & $\checkmark$ & 0.436 & 0.510 & 0.422 & 0.456 \\
        w/o Diversity & $\checkmark$ & $\times$ & $\checkmark$ & 0.393 & 0.472 & 0.428 & 0.431 \\
        w/o Efficiency & $\checkmark$ & $\checkmark$ & $\times$ & 0.401 & 0.468 & \textbf{0.435} & 0.435 \\
        \midrule
        \rowcolor{gray!15}
        \textbf{TopoCurate-SFT (Ours)} & $\checkmark$ & $\checkmark$ & $\checkmark$ & \textbf{0.445} & \textbf{0.529} & \textbf{0.435} & \textbf{0.470} \\
        \bottomrule
    \end{tabular}
    }
    \label{tab:ablation_components}
\end{table}

\begin{table}[t]
    \centering
    \caption{\textbf{Ablation Study (RL).} Impact of topological metrics on Pass@1 accuracy (Qwen3-32B-Thinking as backbone). ``Struct" and ``Div" denote Error Branch Ratio and Strategic Heterogeneity.}
    \resizebox{\linewidth}{!}{
    \begin{tabular}{l|cc|cccc}
        \toprule
        \multirow{2}{*}{\textbf{Method}} & \multicolumn{2}{c|}{\textbf{Components}} & \multicolumn{4}{c}{\textbf{Tau2 Pass@1}} \\
        & \textbf{Struct} & \textbf{Div} & \textbf{Airline} & \textbf{Retail} & \textbf{Telecom} & \textbf{Avg} \\
        \midrule
        TopoCurate-RL (w/o Topology) & $\times$ & $\times$ & 0.531 & 0.643 & 0.721 & 0.632 \\ 
        \midrule
        w/o Structure & $\times$ & $\checkmark$ & 0.542 & 0.661 & 0.748 & 0.650 \\
        w/o Diversity & $\checkmark$ & $\times$ & 0.547 & 0.668 & 0.756 & 0.657 \\
        \midrule
        \rowcolor{gray!15}
        \textbf{TopoCurate-RL (Ours)} & $\checkmark$ & $\checkmark$ & \textbf{0.558} & \textbf{0.687} & \textbf{0.784} & \textbf{0.676} \\
        \bottomrule
    \end{tabular}
    }
    \label{tab:ablation_rl}
    \vspace{-4mm}
\end{table}

\subsection{Complexity and Robustness Analysis.} 
We provide complexity analysis for introduced topology modeling in Appendix \ref{appendix_complex}. We can observe that the extra 10\%-30\% data selection time is clearly less than the basic outcome filtering time. More importantly, the rl task set could also be reduced to 75\% compared with outcome filtering, which will reduce RL training time by 30\% while achieving superior training performance. Moreover, the selection of embedding models and metric weights for data curation are also discussed in Appendix to strengthen the robustness of our method.
\section{Conclusion}
\label{sec:conclusion}
In this work, we introduce TopoCurate, a data curation framework shifting the paradigm from outcome filtering to topological interaction modeling. By projecting linear rollouts into a quotient topology, TopoCurate breaks the outcome equivalence illusion to distill robust behaviors—such as reflective recovery—ignored by standard metrics. Our dual-stage curation mitigates covariate shift in SFT training and maximizes gradient SNR in RL training, establishing new state-of-the-art performance on Tau2 and BFCLv3 benchmarks. This work establishes structure-aware curation as a key driver in building more resilient and adaptable agents.

\section*{Limitations}
We conduct experiments in tool-use settings, including multi-turn function calling and API interaction, which demonstrate the effectiveness of TopoCurate in structured agent-environment scenarios. Nevertheless, its applicability to other agentic settings, such as search, remains to be further explored. Since these scenarios often involve longer contexts, extending embedding-based topology modeling to them may require more careful investigation. We leave this direction to future work.

\section*{Acknowledgements}
This work was supported by the "Pioneer" and "Leading Goose" R\&D Program of Zhejiang (2025C02037), Research Project of the Philosophy and Social Sciences Planning Research Platform (Laboratory) of Zhejiang Province, and Key R\&D Program of Hangzhou (2025SZDA0254).


\bibliography{custom}


\newpage
\appendix

\newpage





\section{More Details about Related Work}
\subsection{Discussion with Structured Rollout Strategy}
Regarding Structured Rollout Strategies, current approaches have developed a highly efficient ``Online Exploration Engine." Techniques such as tree-based sampling and prefix sharing \cite{ji2025tree, li2025treepo, tran2025exploiting, wu2025portool}, along with lookahead branching and critical node search \cite{xing2025lookahead, dihan2025weboperator, shen2025carl,zhong2026sight}, have significantly broadened the scope of exploration and made preliminary strides in addressing the challenge of granular credit assignment in sparse reward settings. However, these methods are fundamentally centered on Online Local Optimization, typically treating tree structures as transient vehicles for immediate gradient estimation or trajectory generation. This focus leaves a critical gap in the Offline Global Valuation of interaction data. As a result, valuable computational resources are often wasted on semantically redundant branches or samples with low signal-to-noise ratios.

In contrast, our TopoCurate framework introduces a novel approach with Offline Topological Analysis. By aggregating and projecting historical interaction trajectories into a unified Semantic Quotient Topology, we can precisely pinpoint high-value structures characterized by reflective recovery from a global perspective in an offline setting, thereby drastically improving data utilization efficiency. Consequently, our method enables a profound synergy with existing online strategies. The frontend leverages Structured Rollout techniques to efficiently generate rich, tree-based candidate trajectories, while the backend employs TopoCurate for offline structural distillation. This ``Online Generation + Offline Selection" paradigm ensures that the model updates its parameters solely based on the most topologically instructive paths, thereby maximizing the marginal benefit of training.

\subsection{Discussion with Environment Scaling}
Current approaches to environment scaling, such as AutoForge \cite{cai2025autoforge}, RandomWorld \cite{sullivan2025procedural}, and AgentScaler \cite{fang2025towards}, have successfully established a powerful expansion engine. By leveraging automated environment synthesis, procedural task generation, and API community detection techniques, these methods have tackled the challenges of limited interaction scenarios and insufficient domain coverage in agent training. However, they are still confined by an outcome-centric filtering paradigm, where SFT trajectories are selected based solely on binary success signals, and RL tasks are chosen based on pass-rate thresholds.

In contrast, our TopoCurate framework fundamentally disrupts this outcome equivalence illusion by introducing topological interaction modeling. This innovation shifts the focus of selection from ``whether it succeeds" to ``how it succeeds." Specifically, TopoCurate projects multi-trial trajectories into a unified Semantic Quotient Topology, constructing Directed Acyclic Graphs (DAGs) to rigorously capture decision bifurcations, error recovery paths, and the causal structure of state transitions—dynamics that existing outcome-based metrics often overlook.

As a result, our method perfectly complements environment scaling pipelines. While expansion techniques provide a vast and varied pool of tasks, TopoCurate performs a structural distillation, ensuring that synthetic data is not ingested blindly. Instead, it is carefully filtered through behavioral dynamics, effectively transforming the benefits of environment scaling into robust decision-making capabilities for agents.

\section{More Details about Experiments}
\subsection{Seed Data Construction and Topological Filtering }
\label{appdendix_data}

This appendix provides implementation details for the seed data construction pipeline.

\paragraph{Multi-Agent Synthesis Framework}:We adapt a multi-agent collaborative framework inspired by APIGen-MT \cite{prabhakar2025apigen} to generate task blueprints and trajectories. The system employs seven specialized agents coordinated by a MasterAgent as follows:

\begin{itemize}[leftmargin=*]
    \item UserProfileAgent: Generates diverse user personas with domain-specific consumption histories.
    \item InstructionAgent: Creates user instructions via two complementary paths:
    \begin{itemize}
        \item \textit{Requirement-First}: Directly generates complex user intents based on target difficulty.
        \item \textit{Tool-Chain-First}: Constructs tool sequences via random walks over API dependency graphs, then converts them to natural language.
    \end{itemize}
    \item EnvironmentAgent: Synthesizes environment states aligned with instruction complexity.
    \item RubricAgent: Defines task success criteria and constraint verification rules.
    \item ValidatorAgent: Performs format, logic, and quality validation.
    \item DeduplicatorAgent: Removes redundancy via semantic embeddings.
\end{itemize}

\paragraph{Data Volume and Domain Distribution}: (i) Airline \& Retail Domains.For Airline and Retail domains, we synthesize 1,000 expert trajectories each for SFT training and 250 tasks for RL training respectively. Tasks are generated using the multi-agent framework and stratified by difficulty levels (simple, moderate, complex);(ii)Telecom Domain. Due to the dual-control nature of Telecom tasks (requiring agent-user coordination for device operations), we do not synthesize tasks from scratch. Instead, we select high-quality Telecom train tasks from the open-source Tau2 official source (different from test task set)  and distill expert trajectories using Claude-4.5-Sonnet for both SFT and RL training.

\paragraph{Pass-Rate-Based Stratification and Our Topological Filtering}: After synthesizing the raw corpus, we stratify tasks based on their empirical pass rates using Qwen3-32B-Instruct (without fine-tuning):(i) SFT Pool (High Pass Rate): Tasks with pass rate $\geq 0.7$ are classified as \textit{fundamental} and allocated to the SFT pool. For these tasks, we distill expert trajectories using Claude-4.5-Sonnet with few-shot prompting and policy constraints. This yields approximately 3000 candidate trajectories (Airline: 1000, Retail: 1000, Telecom: 1000). TopoCurate then applies topological filtering (Sec. \ref{sec:sft}) to select $\sim$2400 high-quality trajectories based on $S_{\text{ref}}$, $S_{\text{eff}}$, and $S_{\text{rare}}$ metrics; (ii) RL Pool (Moderate Pass Rate).Tasks with pass rate $\in [0.1, 0.7]$ exhibit sufficient difficulty and are allocated to the RL pool, yielding approximately 600 candidate tasks. Crucially, the final RL task selection depends on the SFT-trained model's pass rate:We first train an SFT model using the curated SFT data.Then, We re-evaluate the RL task pool using the SFT-trained model to compute updated pass rates $p_{\text{SFT}}(\mathcal{T})$; TopoCurate then applies topological task selection (Sec. \ref{sec:rl}) within the subset $\{\mathcal{T} : p_{\text{SFT}}(\mathcal{T}) \in [0.1, 0.7]\}$ to select $\sim$480 tasks with high structural complexity ($V_{\text{struct}}$, $V_{\text{div}}$).

\paragraph{Quality Assurance}: Each generated sample undergoes three-stage validation:(i) Format Validation: Schema compliance checks;(ii)Logic Validation: Constraint consistency verification; (iii) Quality Validation: Semantic coherence assessment;(iv) Semantic deduplication is performed using embedding-based similarity matching (threshold: cosine similarity $> 0.92$), reducing redundancy by approximately 12\%.


\subsection{Training Configurations}

\paragraph{Supervised Fine-Tuning Configuration.}
All experiments are conducted on the Tau2 benchmark environment with mixed-domain training across Airline, Retail, and Telecom tasks. For SFT, we fine-tune the Qwen3-Thinking models using a global batch size of 256. The learning rate is selected from the range {5e-6, 7e-6} via grid search on the validation set. We train for 3 epochs using the AdamW optimizer with weight decay 0.01. The learning rate schedule employs linear warmup for 10\% of total steps followed by cosine decay \cite{wang2025scaling,chen2025revisiting}. 
The mixed-domain training strategy ensures the model learns generalizable tool-use patterns across diverse task distributions.

\paragraph{Reinforcement Learning Configuration.}
For RL training, we apply Group Relative Policy Optimization (GRPO) on the SFT-trained model using a rollout batch size of 1,024. Each rollout iteration samples 60 tasks in total, with 20 tasks from each domain (Airline, Retail, and Telecom) to maintain balanced exploration. For each sampled task, we generate 16 trajectories to compute group-relative advantages. The reward function is binary sparse: $r=1$ for task success and $r=0$ otherwise. We use the AdamW optimizer with a learning rate of 1e-6 and train for 80 steps(8 epoch), at which point convergence is observed based on training reward curves. 


\subsection{Embedding Model Selection for Topology Modeling}
\label{appdendix_embedding}

 We conducted ablation experiments with two open-source embedding models of varying scales across 99 tasks equally sampled from Tau2 domains. Through detailed embedding and statistical analysis of the merging results. Take Jina as the baseline, we can observe that all three models show high agreement on the Tier construction (merging feature) and the final decision of "which trajectories should be selected as training data". These results can consistently verify the robustness of our method.

\begin{table}[htbp]
    \centering
    \caption{\textbf{Embedding Model Selection.} Comparison of different embedding models used in the Tier construction phase.\textsuperscript{$\dagger$} represents the correlation coefficient between the selected trajectories and the ground truth ranking.}
    \resizebox{0.45\textwidth}{!}{
    \begin{tabular}{l|cccccc}
        \toprule
        \textbf{Model} & \textbf{Params} & \textbf{Dim} & \textbf{Context} & \textbf{Tier Agreement} & \textbf{Spearman $\rho$\textsuperscript{$\dagger$}} \\
        \midrule
        Jina-v2-base-en & 137M & 768 & 8,192 & --- & --- \\
        \midrule
        Owen3-Embedding-0.6B & 600M & 1,024 & 32,768 & 87.8\% & 0.945 \\
        \bottomrule
    \end{tabular}
    }
    \label{tab:embedding_ablation}
    \smallskip
\end{table}

\subsection{Sensitivity Analysis of  Metric Weights}
\label{app:hyperparams}

During the trajectory selection phase, we conducted a systematic sensitivity analysis on the weights of the selection metrics. As presented in Table~\ref{tab:weight_ablation}, where our default weight configuration is $\alpha=0.4, \beta=0.3, \gamma=0.3$, we introduced perturbations to the weights to evaluate performance stability. The comparison reveals that the set of selected trajectories and key statistical metrics remained largely stable across different settings. The patterns of trajectory retention/exclusion and the overall selection outcomes exhibited only minor fluctuations without affecting the primary findings. These results demonstrate that the proposed trajectory selection method exhibits strong robustness to weight settings and does not depend on fine-grained weight tuning to achieve stable performance.

\begin{table}[htbp]
    \centering
    \caption{\textbf{Sensitivity Analysis of Trajectory Selection Metric Weights.} The weight settings are denoted as $(\alpha, \beta, \gamma)$, corresponding to the weights for Efficiency, Diversity, and Reflection metrics, respectively.}
    \resizebox{0.45\textwidth}{!}{
    \begin{tabular}{l|cccc}
        \toprule
        \textbf{Weight Settings $(\alpha, \beta, \gamma)$} & \textbf{Airline} & \textbf{Retail} & \textbf{Telecom} & \textbf{Avg} \\
        \midrule
        $(0.4, 0.3, 0.3)$ & 0.445 & 0.529 & 0.435 &  0.470\\ 
        $(0.4, 0.2, 0.4)$ & 0.413 & 0.549 & 0.420 &  0.461\\
        $(0.3, 0.4, 0.4)$ & 0.415 & 0.517 & 0.424 &  0.452\\
        \bottomrule
    \end{tabular}
    }
    \label{tab:weight_ablation}
\end{table}

\begin{figure*}[t]
    \centering
    \begin{subfigure}[b]{0.32\textwidth}
        \centering
        \includegraphics[width=\linewidth]{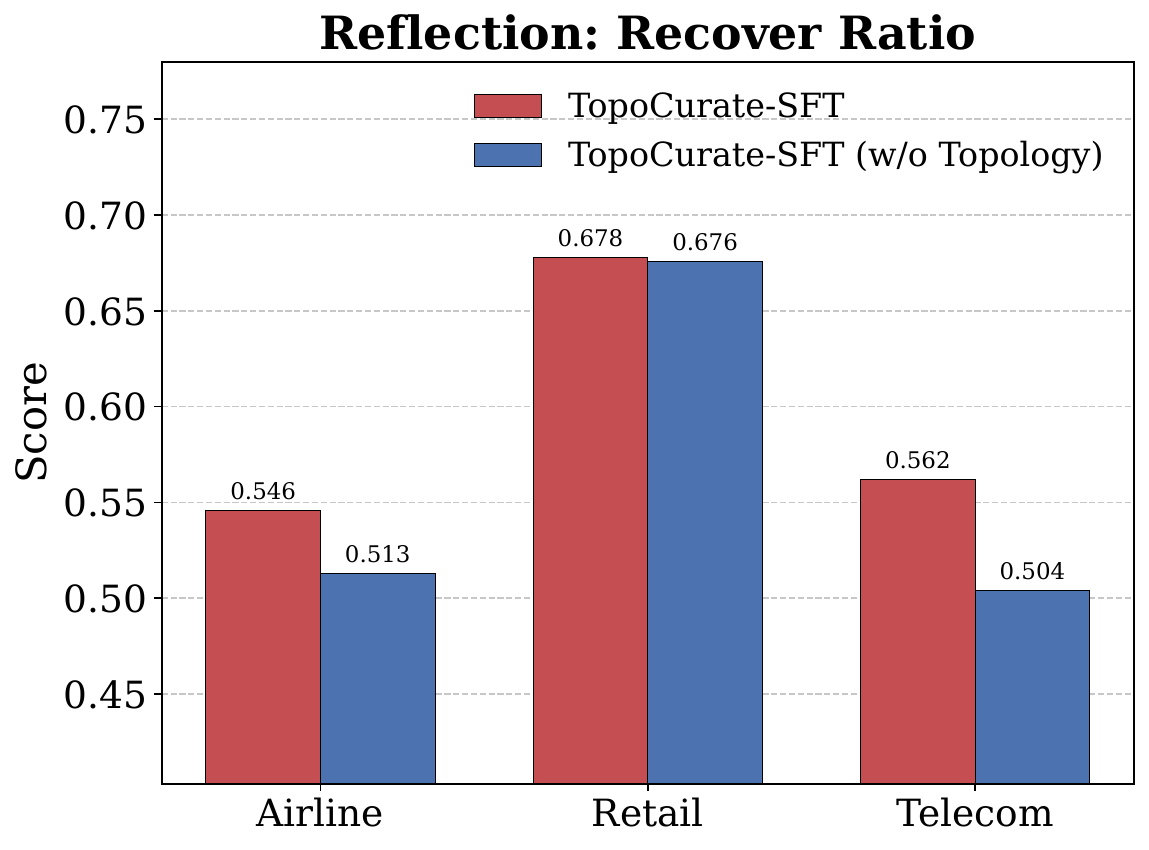}
        \caption{\textbf{Reflection} }
        \label{fig:analysis_reflection}
    \end{subfigure}
    \hfill
    \begin{subfigure}[b]{0.32\textwidth}
        \centering
        \includegraphics[width=\linewidth]{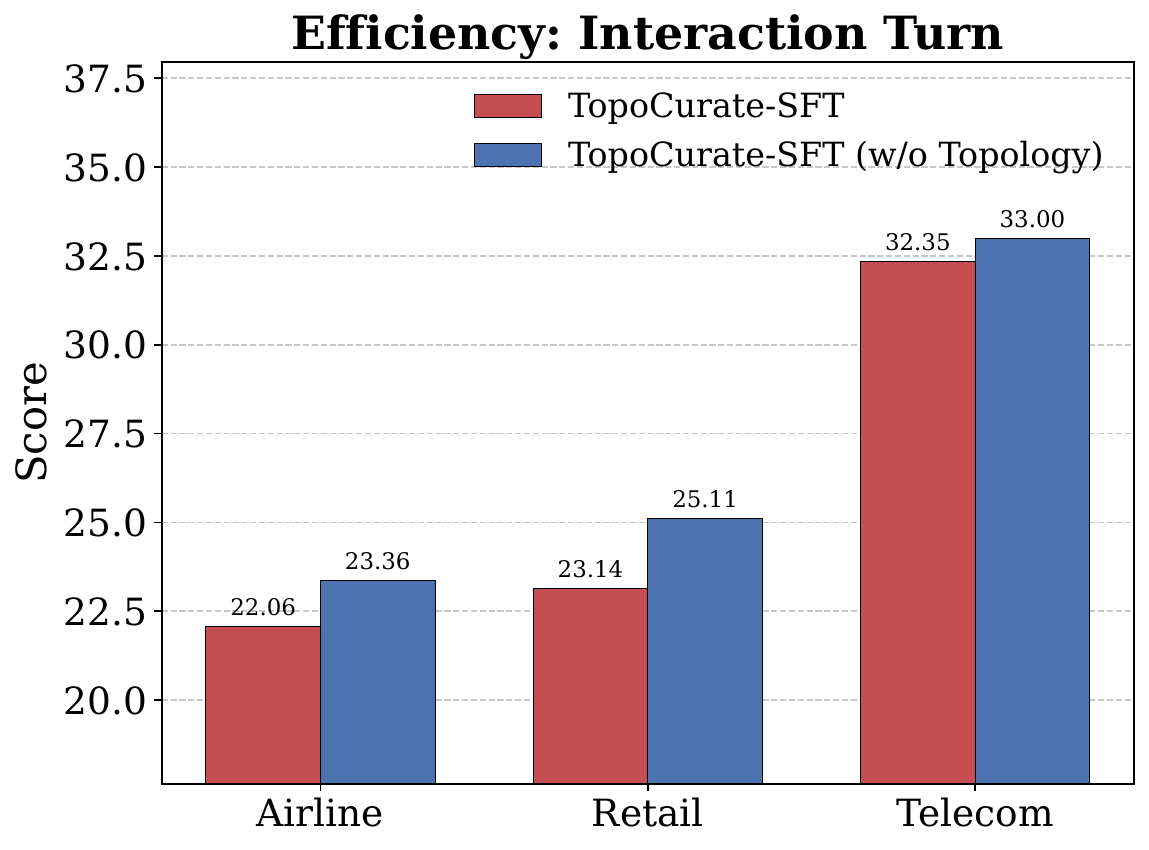}
        \caption{\textbf{Efficiency} }
        \label{fig:analysis_efficiency}
    \end{subfigure}
    \hfill
    \begin{subfigure}[b]{0.32\textwidth}
        \centering
        \includegraphics[width=\linewidth]{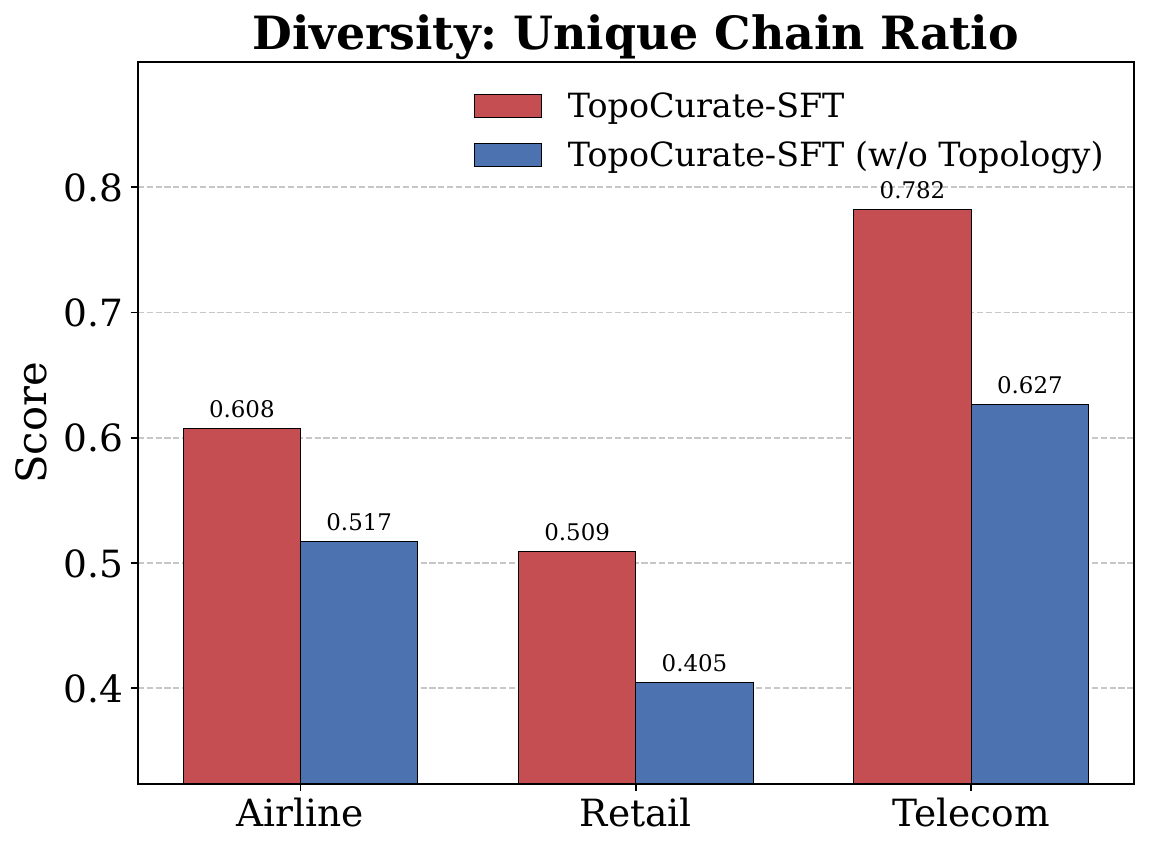}
        \caption{\textbf{Diversity} }
        \label{fig:analysis_diversity}
    \end{subfigure}
   \caption{\textbf{Model Behavior Analysis.} We compare the behavioral patterns of TopoCurate-SFT (Ours) vs. the baseline SFT (w/o Topology). \textbf{(a)} Ours demonstrates significantly higher reflective recovery rates, particularly in the complex Telecom domain. \textbf{(b)} In terms of the number of interaction turns (lower is better), Ours achieves higher efficiency overall, while maintaining success rate across all domains. \textbf{(c)} Ours consistently generates more diverse tool chains (Unique Chain Ratio), mitigating mode collapse.}
    \label{fig:behavior_analysis}
\end{figure*}
\subsection{RL Training Dynamics.} 
\begin{figure*}[t]
    \centering
    \includegraphics[width=0.48\linewidth]{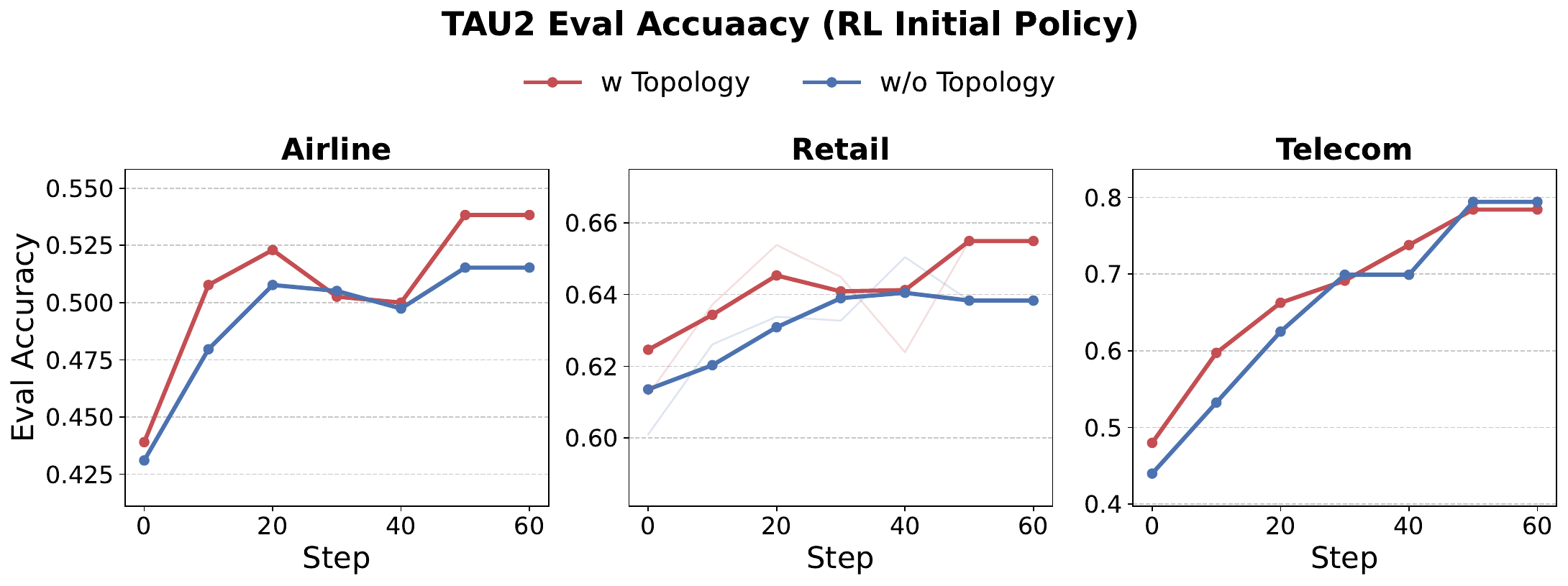}
    \includegraphics[width=0.48\linewidth]{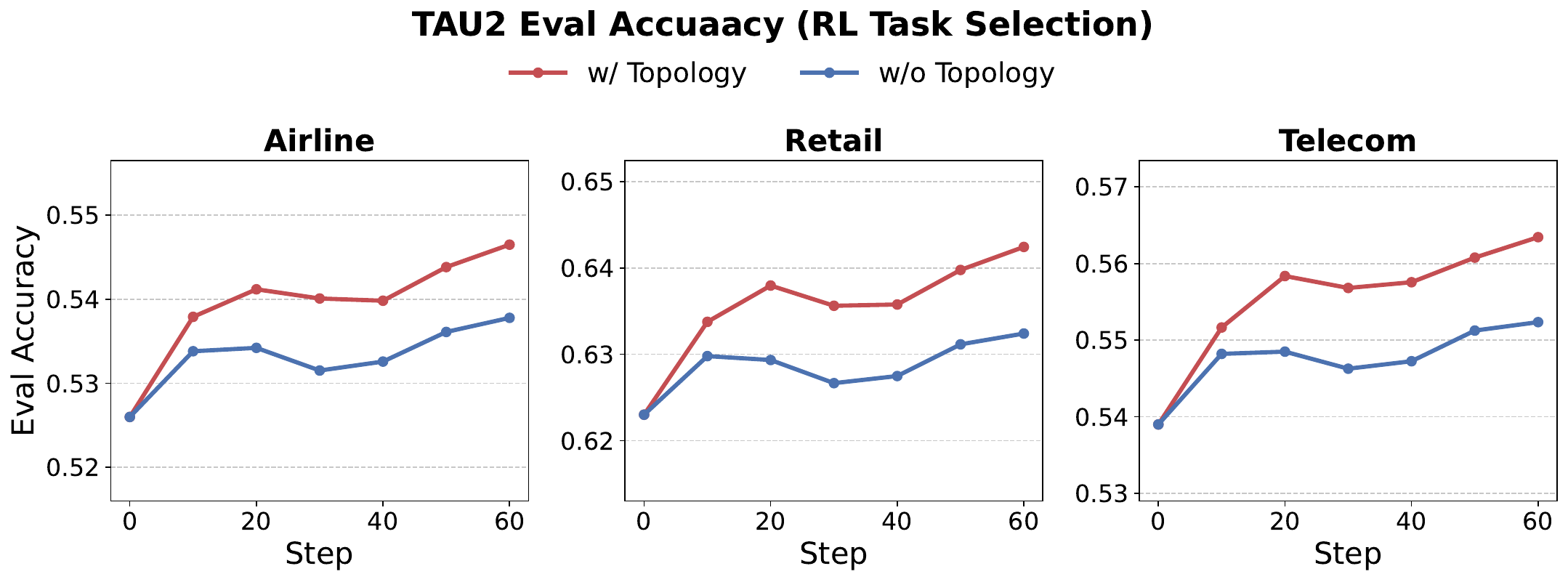}
    \caption{\textbf{Evaluation Accuracy on Tau2 for RL.} (Left) Impact of SFT initialization quality on downstream RL performance across Airline, Retail, and Telecom domains. Models initialized with topological SFT exhibit consistently higher accuracy. (Right) Impact of RL task selection on final performance. Topologically-selected tasks lead to superior convergence across all domains, with the most pronounced gains in Telecom.}
    \label{fig:rl_eval_accuracy}
\end{figure*}

\begin{figure}[t]
    \centering
    \includegraphics[width=0.45\linewidth]{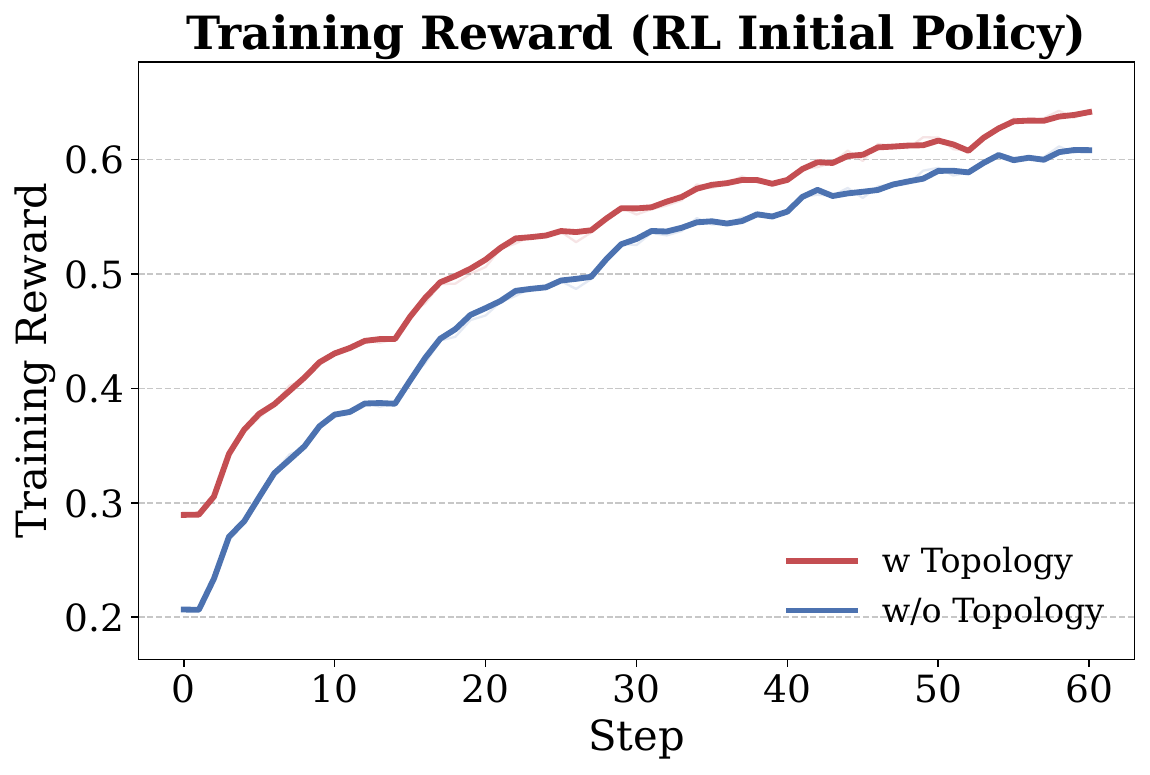}
    \includegraphics[width=0.45\linewidth]{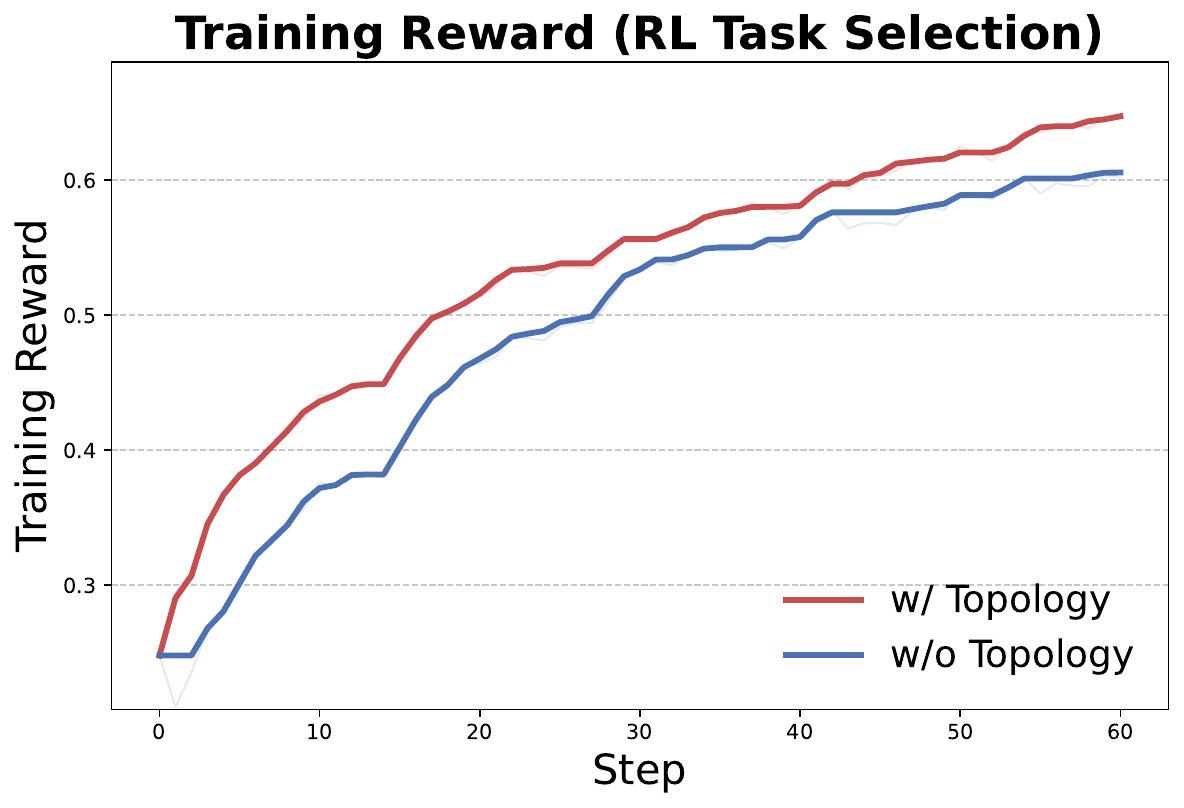}
    \caption{\textbf{Training Reward Curves for RL.} (Left) Comparison of models initialized with topologically-curated SFT (w/ Topology) vs. outcome-filtered SFT (w/o Topology), using the same RL task pool. (Right) Comparison of models trained on topologically-selected RL tasks (w/ Topology) vs. uniformly-sampled tasks (w/o Topology) from the same SFT checkpoint. Topological curation improves both initialization quality and task selection efficiency.}
    \label{fig:rl_training_reward}
\end{figure}

To validate topological curation's effectiveness across the training pipeline, we conduct two controlled experiments on Qwen3-32B-Thinking, isolating the impact of (i) SFT initialization and (ii) RL task selection. \textbf{Impact of SFT Initialization Quality.}
We compare two RL runs starting from different SFT checkpoints: one trained on TopoCurate-selected trajectories (w/ Topology) and another on outcome-filtered data (w/o Topology), using the same RL task pool. As shown in Figure \ref{fig:rl_training_reward} (left), the topologically-initialized model exhibits significantly higher starting training reward (0.29 vs. 0.22 at step 0) and converges to a superior plateau (0.64 vs. 0.61 at step 60), demonstrating that high-quality SFT provides a stronger foundation for policy optimization. Evaluation accuracy on Tau2 (Figure \ref{fig:rl_eval_accuracy}, left) corroborates this: the w/ Topology model consistently outperforms the baseline across all domain;\textbf{Impact of RL Task Selection.}
To isolate the effect of task selection, we train two models from the same SFT checkpoint (ensuring identical initial training rewards of $\sim$0.25) but vary the RL task pool: one selected via topological metrics ($V_{\text{struct}}, V_{\text{div}}$) and another via uniform sampling. Figure \ref{fig:rl_training_reward} (right) reveals that while both start identically, topologically-selected tasks yield steeper learning curves and higher final rewards . Evaluation accuracy (Figure \ref{fig:rl_eval_accuracy}, right) confirms this: the w/ Topology model achieves consistent gains across all domains, with Telecom showing substantial improvement , validating that topological task selection maximizes gradient SNR.

These experiments establish a causal chain: topological SFT curation $\to$ robust initialization $\to$ improved RL starting point; topological task selection $\to$ high-SNR gradients $\to$ accelerated convergence. Together, they demonstrate TopoCurate's compositional impact—both stages contribute independently and synergistically to final performance.

\subsection{More Evaluation on BFCLv4}\label{appendix_bfcl}
We further evaluated on BFCL v4 as shown in Table \ref{tab:bfcl_v4_results}, a comprehensive benchmark covering single/multi-turn function calling, live API invocation, agentic memory tasks, and hallucination detection. Results on Qwen3-32B can further verify the effectiveness of our method.

\begin{table}[htbp]
    \centering
    \caption{\textbf{BFCL v4 Results on Qwen3-32B.} Evaluation across multiple sub-tasks. The best result in each column is highlighted in bold.}
    \label{tab:bfcl_v4_results}
    \resizebox{0.45\textwidth}{!}{ 
    \begin{tabular}{lccccc}
        \toprule
        \textbf{Method} & \textbf{NON\_LIVE} & \textbf{LIVE} & \textbf{MULTI\_TURN} & \textbf{AGENTIC} & \textbf{HALLUCINATION} \\
        \midrule
        Qwen3-32B-Thinking (Base) & 0.475 & 0.700 & 0.215 & 0.197 & 0.724 \\
        TopoCurate-SFT (w/o Topology) & 0.531 & 0.799 & 0.253 & 0.201 & 0.703 \\
        TopoCurate-SFT & 0.550 & 0.790 & 0.261 & 0.227 & 0.713 \\
        TopoCurate-RL (w/o Topology) & 0.539 & 0.798 & 0.301 & 0.255 & 0.783 \\
        \textbf{TopoCurate-RL} & \textbf{0.562} & \textbf{0.805} & \textbf{0.325} & \textbf{0.261} & \textbf{0.803} \\
        \bottomrule
    \end{tabular}
    }
\end{table}

\subsection{Topology Construction Statistics.}
We agree that detailed topology statistics are essential for assessing whether the quotient topology is dense enough to support reliable Success Potential estimation. Table~\ref{tab:topology_stats} reports the construction statistics across all three Tau2 domains, computed over the full RL candidate pool with 16 rollouts per task and default thresholds $\delta_{\mathrm{tool}}=0.92$ and $\delta_{\mathrm{result}}=0.88$.

\begin{table}[htbp]
    \centering
    \caption{\textbf{Topology construction statistics across Tau2 domains.} We report the average number of rollouts per task, raw turns before merging, merged nodes after quotient construction, and the resulting compression ratio.}
    \label{tab:topology_stats}
    \resizebox{0.45\textwidth}{!}{
    \begin{tabular}{lcccc}
        \toprule
        \textbf{Domain} & \textbf{Avg Rollouts/Task} & \textbf{Raw Turns (Mean)} & \textbf{Merged Nodes (Mean)} & \textbf{Compression Ratio} \\
        \midrule
        Airline & 16 & 392 & 87  & 4.5$\times$ \\
        Retail  & 16 & 428 & 102 & 4.2$\times$ \\
        Telecom & 16 & 526 & 138 & 3.8$\times$ \\
        \bottomrule
    \end{tabular}
    }
\end{table}

The average number of rollouts per task is set to $K=16$, establishing the baseline sample volume to ensure sufficient coverage of diverse decision pathways, including both success and failure branches. The raw turns reflect the cumulative number of interaction steps across all rollouts before merging, and the relatively high counts (392--526) highlight the long-horizon complexity of the multi-turn environments. The merged nodes denote the cardinality of the unique state space after applying the semantic equivalence relation in Eq.~(3), where each node represents a semantically distinct logical state, such as a specific API milestone or environmental response. The compression ratio, computed as raw turns divided by merged nodes, ranges from 3.8$\times$ to 4.5$\times$, indicating that the quotient topology removes redundant behaviors while preserving critical, parameter-sensitive decision states. This balance supports reliable estimation of the Success Potential Field $\Phi(v)$.

These statistics confirm that the topology achieves sufficient densification for $\Phi(v)$ estimation without over-merging semantically distinct states. The conservative compression ratios reflect the dual-constraint design in Eq.~(3), which preserves important distinctions such as different API parameters. More importantly, the visitation statistics show a clear practical pattern: a subset of semantically central nodes, such as user verification, product search, or order checkout, is revisited multiple times across different rollouts and thus accumulates sufficient empirical evidence. Meanwhile, sparse leaf nodes remain low-count and are handled conservatively in downstream scoring, primarily contributing to the diversity-oriented term $S_{\mathrm{rare}}$. This is the regime in which empirical Success Potential estimation becomes meaningful. We will add this table and the accompanying visitation analysis to Section~5.5 in the revision.

\subsection{Comparison with Curriculum RL / DUMP.}
TopoCurate is a structural refinement on top of pass-rate selection and is complementary to curricula such as DUMP. Our RL pool is already restricted to the medium-difficulty band $p_{\mathrm{SFT}} \in [0.1, 0.7]$ (Appendix~B.1), which is the regime curriculum methods typically target, and TopoCurate adds structural filtering within this band.

The key motivation is the \emph{Outcome Equivalence Illusion}: tasks with the same pass rate or advantage magnitude can still have very different training value, since advantage is only a scalar variance signal and cannot capture genuine bifurcation or solution-path diversity. To validate this, we added a DUMP-style advantage-UCB baseline. Table~\ref{tab:dump_comparison} reports the Tau2 Pass@1 results on Qwen3-32B.

\begin{table}[htbp]
    \centering
    \caption{\textbf{Comparison with curriculum RL / DUMP on Qwen3-32B (Tau2 Pass@1).} The best result in each column is highlighted in bold.}
    \label{tab:dump_comparison}
    \resizebox{0.45\textwidth}{!}{
    \begin{tabular}{lcccc}
        \toprule
        \textbf{Method (RL task selection)} & \textbf{Airline} & \textbf{Retail} & \textbf{Telecom} & \textbf{Avg} \\
        \midrule
        w/o Topology (uniform / pass-rate) & 0.531 & 0.643 & 0.721 & 0.632 \\
        DUMP (advantage-UCB curriculum)    & 0.540 & 0.656 & 0.744 & 0.647 \\
        \textbf{TopoCurate-RL (Ours)}      & \textbf{0.558} & \textbf{0.687} & \textbf{0.784} & \textbf{0.676} \\
        \bottomrule
    \end{tabular}
    }
\end{table}

TopoCurate-RL achieves 0.676 versus 0.647 for DUMP, showing that structural selection adds gains beyond difficulty-based curriculum alone. This result supports the claim that topology-aware filtering provides additional benefits beyond curriculum learning based only on task difficulty.

\subsection{Robustness to Extremely Low Selection Ratios.}
We further evaluated TopoCurate under extremely low selection ratios of 5\%, 10\%, and 20\% for SFT on Qwen3-8B. For a fair comparison, we keep the amount of training data the same between the Outcome-based method and TopoCurate on the Tau2 benchmark. Table~\ref{tab:low_ratio} summarizes the results.

\begin{table}[htbp]
    \centering
    \caption{\textbf{Robustness of TopoCurate under low selection ratios on Qwen3-8B.} The best result in each row is highlighted in bold.}
    \label{tab:low_ratio}
    \resizebox{0.48\textwidth}{!}{
    \begin{tabular}{lccc}
        \toprule
        \textbf{Selection Ratio} & \textbf{Outcome-based} & \textbf{TopoCurate-SFT} & \textbf{Relative Gain} \\
        \midrule
        5\%  & 0.312 & 0.389 & +24.7\% \\
        10\% & 0.368 & 0.421 & +14.4\% \\
        20\% & 0.405 & 0.448 & +10.6\% \\
        \bottomrule
    \end{tabular}
    }
\end{table}

The reported results verify that TopoCurate consistently outperforms the Outcome-based method across all tested selection ratios. This demonstrates that TopoCurate remains robust even when only a small fraction of the data is retained. We will add this analysis to Section~5.3 in the revision.

\section{More Details about Method}
\label{app:theory}
This appendix provides formal theoretical justifications for TopoCurate's trajectory and task selection mechanisms, grounding our empirical design choices in learning-theoretic principles.

\subsection{Computational Complexity of Topology Construction.}\label{appendix_complex}
Constructing the quotient topology involves pairwise comparisons of interaction turns. A naive implementation scales quadratically $O(N_{\text{turn}}^2)$. To ensure scalability, we employ \textit{Locality-Sensitive Hashing (LSH)} on the multi-view embeddings to retrieve candidate equivalent turns, reducing the complexity to $O(N_{\text{turn}} \log N_{\text{turn}})$. Furthermore, since topological curation is an offline pre-processing step, it incurs zero computational overhead during the actual training loop.

\paragraph{Time Cost Analysis.} We acknowledge that constructing a semantic quotient topology will introduce an additional processing stage compared with standard outcome-based filtering. \textbf{But the extra time cost during data selection could be acceptable or meaningful in some way as the following table shows.}

\begin{table}[htbp]
    \centering
    \caption{Time consumption comparison for topology construction and filtering on three domains.}
    \label{tab:time_consumption}
    \resizebox{0.45\textwidth}{!}{ 
    \begin{tabular}{lccc}
        \toprule
         & \textbf{Telecom} & \textbf{Retail} & \textbf{Airline} \\
        \midrule
        Construction\_Topology & 38.3 min & 33.3 min & 40.1 min \\
        Outcome\_Reward\_Filtering & 300.5 min & 233.6 min & 227.2 min \\
        \bottomrule
    \end{tabular}
    }
\end{table}

Take the task selection for rl training for example, previous pass-rate-selected (outcome-based filtering) strategy also needs the process of generating multi-trial rollouts, while we just further provide a simple but effective topology analysis to evaluate the training potential of training tasks. We just introduce extra 10\%-30\% data selection time, the rl task set could also be reduced to 75\% compared with outcome filtering, which means will reduces RL training time by 30\% while achieving superior training performance (6.9\% performance gain).

\subsection{Mitigate the Covariate Shift and Mode Collaps through Trajectory Selection}
\label{app:theory_sft}

Our mechanism can be viewed as a generalized Rejection Sampling scheme. While standard Outcome Filtering assumes a uniform prior ($p(\tau) \propto \mathbb{I}[R=1]$), TopoCurate modulates this prior using the topological potential:
\begin{equation}
    p(\tau) \propto \mathbb{I}[R=1] \cdot \exp(w(\tau))
\end{equation}
This effectively shifts the sampling distribution towards the process quality, suppressing suboptimal lucky guesses. Formally, SFT minimizes the negative log-likelihood $\mathcal{L}_{\text{SFT}} = -\mathbb{E}_{\tau \sim p(\tau)}[\log \pi_\theta(\tau)]$. By reweighting the data distribution, our method explicitly minimizes the KL divergence $D_{\text{KL}}(\pi_{\theta} \| \pi_{\text{expert}})$ between the learned policy and an ideal robust expert that balances recovery, efficiency, and diversity. This realignment mitigates Covariate Shift (by including error-recovery patterns) and Mode Collapse (by enforcing diversity), addressing the dual deficits of outcome filtering.

\paragraph{Problem Formulation: SFT as Distribution Matching.}
Supervised Fine-Tuning minimizes the forward KL divergence between the policy $\pi_\theta$ and the empirical data distribution $\hat{p}_{\text{data}}$:
\begin{equation}
\begin{aligned}
\mathcal{L}_{\mathrm{SFT}}(\theta)
&= \mathbb{E}_{\tau \sim \hat{p}_{\mathrm{data}}}
   \bigl[-\log \pi_\theta(\tau)\bigr] \\
&= D_{\mathrm{KL}}(\hat{p}_{\mathrm{data}} \| \pi_\theta)
   + \mathrm{const.}
\end{aligned}
\end{equation}
Standard outcome filtering defines $\hat{p}_{\text{data}}(\tau) \propto \mathbb{I}[R(\tau)=1]$, implicitly assuming all successful trajectories are equally valuable. This uniform weighting ignores critical distinctions in process quality, leading to three pathologies:

\paragraph{Pathology 1: Covariate Shift from Fragile Trajectories.}
Consider a task with two solution paths: $\tau_{\text{fragile}}$ succeeds via a lucky sequence that avoids constraint checking, while $\tau_{\text{robust}}$ explicitly validates constraints before proceeding. Standard filtering assigns equal weight: $\hat{p}(\tau_{\text{fragile}}) = \hat{p}(\tau_{\text{robust}})$. However, at test time, environment perturbations (e.g., stricter error handling) render $\tau_{\text{fragile}}$ invalid, causing distribution shift. Formally, let $p_{\text{test}}$ denote the test distribution. The expected test loss suffers from:
\begin{equation}
\begin{aligned}
\mathbb{E}_{\tau \sim p_{\mathrm{test}}}[\ell(\pi_\theta, \tau)]
&\ge \mathbb{E}_{\tau \sim \hat{p}_{\mathrm{data}}}[\ell(\pi_\theta, \tau)] \\
&\quad + D_{\mathrm{TV}}(p_{\mathrm{test}}, \hat{p}_{\mathrm{data}})
\end{aligned}
\end{equation}
where $D_{\text{TV}}$ is the total variation distance. By prioritizing $\tau_{\text{robust}}$ via Reflective Recovery ($S_{\text{ref}}$), TopoCurate explicitly reduces $D_{\text{TV}}(p_{\text{test}}, \hat{p}_{\text{data}})$ by aligning the training distribution with robust behaviors that generalize under perturbations.

\paragraph{Pathology 2: Token Inefficiency from Redundant Loops.}
Let $\tau_{\text{verbose}}$ contain redundant API calls (e.g., querying the same flight ID thrice) while $\tau_{\text{concise}}$ achieves the same outcome efficiently. Standard filtering weighs them equally, but inference cost scales linearly with trajectory length. The expected inference cost is:
\begin{equation}
C_{\text{infer}} = \mathbb{E}_{\tau \sim \hat{p}_{\text{data}}}[|\tau|] \cdot c_{\text{token}}
\end{equation}
where $c_{\text{token}}$ is the per-token cost. By penalizing $\tau_{\text{verbose}}$ via Semantic Efficiency ($S_{\text{eff}}$), TopoCurate minimizes $\mathbb{E}[|\tau|]$, directly optimizing for token economy. Empirically, this reduces inference latency by up to 20\% (see Sec. \ref{sec:ablation}).

\paragraph{Pathology 3: Mode Collapse from Solution Homogeneity.}
Suppose 80\% of successful trajectories follow a dominant path $\tau_{\text{common}}$, while 20\% explore alternative strategies $\{\tau_{\text{rare},i}\}$. Uniform weighting biases the policy toward $\tau_{\text{common}}$, suppressing exploration. This manifests as mode collapse: the policy fails on tasks requiring $\tau_{\text{rare}}$. Formally, the entropy of the learned policy is bounded by:
\begin{equation}
H(\pi_\theta) \leq H(\hat{p}_{\text{data}}) + \log N_{\text{support}}
\end{equation}
where $N_{\text{support}}$ is the effective support size of $\hat{p}_{\text{data}}$. By upweighting $\tau_{\text{rare}}$ via Distributional Diversity ($S_{\text{rare}}$), TopoCurate increases $N_{\text{support}}$, maximizing $H(\pi_\theta)$ and mitigating mode collapse.

\paragraph{Optimality of Weighted Sampling.}
TopoCurate reweights the data distribution as:
\begin{equation}
\hat{p}_{\text{TopoCurate}}(\tau) \propto \mathbb{I}[R(\tau)=1] \cdot \exp(w(\tau))
\end{equation}
where $w(\tau) = \lambda_{\text{ref}} S_{\text{ref}}(\tau) + \lambda_{\text{eff}} S_{\text{eff}}(\tau) + \lambda_{\text{rare}} S_{\text{rare}}(\tau)$. This can be viewed as tilted importance sampling with target distribution $p^*(\tau) \propto p_{\text{expert}}(\tau)$, where $p_{\text{expert}}$ is a hypothetical robust expert policy. The KL divergence between the learned policy and the expert is:
\begin{equation}
\begin{aligned}
D_{\mathrm{KL}}(\pi_\theta \| p_{\mathrm{expert}})
&= \mathbb{E}_{\tau \sim p_{\mathrm{expert}}}
   \bigl[\log p_{\mathrm{expert}}(\tau)\bigr] \\
&\quad - \mathbb{E}_{\tau \sim p_{\mathrm{expert}}}
   \bigl[\log \pi_\theta(\tau)\bigr]
\end{aligned}
\end{equation}
By aligning $\hat{p}_{\text{TopoCurate}}$ with $p_{\text{expert}}$, TopoCurate directly minimizes this divergence, ensuring the policy internalizes robust, efficient, and diverse strategies.

\subsection{Maximizing Signal-to-Noise Ratio through Task Selection}
\label{app:theory_rl}
In GRPO, the policy gradient's informativeness is determined by the variance of advantages $\text{Var}[\hat{A}]$. Tasks yielding homogeneous rollouts (all succeed or all fail) produce $\text{Var}[\hat{A}] \approx 0$, resulting in vanishing gradient updates. Our structural metrics $V_{\text{struct}}$ and $V_{\text{div}}$ serve as tractable proxies for the Signal-to-Noise Ratio (SNR) of the gradient: tasks with high $V_{\text{struct}}$ exhibit sharp decision boundaries, leading to large advantage gaps and high-variance gradients. Formally, the Fisher information matrix $\mathcal{I}(\theta) = \mathbb{E}[\nabla \log \pi \nabla \log \pi^\top]$ quantifies the sensitivity of the policy to parameter changes. Tasks with high structural variance $\mathbb{E}[\text{Var}(\Phi)]$ maximize $\mathcal{I}(\theta)$, ensuring each gradient update carries maximal information about the optimal policy. Thus, our task selection directly maximizes gradient SNR by prioritizing tasks where outcomes significantly diverge based on critical actions, effectively addressing the fundamental vanishing signal problem in sparse-reward settings.

\paragraph{Problem Formulation: RL Gradient Informativeness.}
In sparse-reward RL with Group Relative Policy Optimization (GRPO), the policy gradient is:
\begin{equation}
\begin{aligned}
\nabla_\theta J(\theta)
&= \mathbb{E}_{\mathcal{T}\sim P_{\mathrm{task}}}
   \mathbb{E}_{\{\tau_k\}_{k=1}^K\sim\pi_\theta(\cdot\mid\mathcal{T})}
   \Bigl[ \\
&\qquad \sum_{k=1}^K
   \nabla_\theta \log \pi_\theta(\tau_k)
   \cdot \hat{A}(\tau_k)
   \Bigr]
\end{aligned}
\end{equation}
where $\hat{A}(\tau_k) = r(\tau_k) - \bar{r}$ is the group-relative advantage. The gradient's Signal-to-Noise Ratio (SNR) is:
\begin{equation}
\text{SNR} = \frac{\|\mathbb{E}[\nabla_\theta J]\|^2}{\text{Var}[\nabla_\theta J]}
\end{equation}
Standard task selection (e.g., uniform sampling or pass-rate thresholds) fails to account for the structural properties of tasks that determine SNR, leading to two failure modes:

\paragraph{Failure Mode 1: Vanishing Gradients from Homogeneous Rollouts.}
Consider a task $\mathcal{T}_{\text{trivial}}$ where all $K$ rollouts yield identical outcomes (either all succeed or all fail). The advantage variance is:
\begin{equation}
\text{Var}[\hat{A}] = \frac{1}{K} \sum_{k=1}^K (\hat{A}(\tau_k) - 0)^2 = 0
\end{equation}
since $\hat{A}(\tau_k) = r - \bar{r} = 0$ for all $k$. Consequently, $\nabla_\theta J \approx 0$, providing no learning signal. The expected gradient magnitude is bounded by:
\begin{equation}
\|\mathbb{E}[\nabla_\theta J]\| \leq \|\nabla_\theta \log \pi\| \cdot \text{Var}[\hat{A}]^{1/2}
\end{equation}
For tasks with $\text{Var}[\hat{A}] \approx 0$, the gradient vanishes regardless of the policy's capacity. TopoCurate avoids such tasks by prioritizing high Error Branch Ratio ($V_{\text{struct}}$), ensuring $\text{Var}[\hat{A}] \gg 0$.

\paragraph{Failure Mode 2: Low Gradient Variance from Mode Collapse.}
Consider a task $\mathcal{T}_{\text{single-path}}$ solvable via only one tool sequence. All successful rollouts follow the same path, leading to low policy entropy:
\begin{equation}
H(\pi_\theta | \mathcal{T}) = -\sum_{\tau} \pi_\theta(\tau|\mathcal{T}) \log \pi_\theta(\tau|\mathcal{T}) \to 0
\end{equation}
Low entropy implies the policy has converged prematurely to a single mode, suppressing exploration. The Fisher information matrix $\mathcal{I}(\theta)$ quantifies the sensitivity of the policy:
\begin{equation}
\mathcal{I}(\theta) = \mathbb{E}_{\tau \sim \pi_\theta}[\nabla_\theta \log \pi_\theta(\tau) \nabla_\theta \log \pi_\theta(\tau)^\top]
\end{equation}
For single-mode distributions, $\mathcal{I}(\theta)$ has low rank, limiting gradient diversity. TopoCurate selects tasks with high Strategic Heterogeneity ($V_{\text{div}}$), ensuring $\mathcal{I}(\theta)$ remains full-rank and gradient updates explore diverse directions.

\paragraph{Optimality of Structural Metrics.}
TopoCurate defines the task sampling distribution as:
\begin{equation}
P_{\text{select}}(\mathcal{T}) \propto \exp\left( \frac{V_{\text{struct}}(\mathcal{T}) + \alpha V_{\text{div}}(\mathcal{T})}{T} \right)
\end{equation}
This prioritizes tasks where:
\begin{itemize}
\item High $V_{\text{struct}}$ ensures sharp decision boundaries, maximizing $\text{Var}[\hat{A}]$ and gradient magnitude.
\item High $V_{\text{div}}$ ensures multiple valid solution paths, maximizing $\text{rank}(\mathcal{I}(\theta))$ and preventing premature convergence.
\end{itemize}
The expected SNR under TopoCurate's distribution is:
\begin{equation}
\mathbb{E}_{\mathcal{T} \sim P_{\text{select}}}[\text{SNR}] \propto \mathbb{E}[V_{\text{struct}}] \cdot \mathbb{E}[V_{\text{div}}]
\end{equation}
Maximizing this expectation directly optimizes for gradient informativeness.

\paragraph{Connection to Fisher Information and Natural Gradients.}
The Fisher information matrix $\mathcal{I}(\theta)$ defines the natural gradient:
\begin{equation}
\tilde{\nabla}_\theta J = \mathcal{I}(\theta)^{-1} \nabla_\theta J
\end{equation}
Natural gradients converge faster than standard gradients by accounting for the curvature of the policy manifold. Tasks with high $V_{\text{struct}}$ and $V_{\text{div}}$ increase the smallest eigenvalue $\lambda_{\min}(\mathcal{I}(\theta))$, reducing the condition number $\kappa = \lambda_{\max}/\lambda_{\min}$ and accelerating convergence. Formally:
\begin{equation}
\|\theta^{(t+1)} - \theta^*\| \leq \left(1 - \frac{\eta}{\kappa}\right) \|\theta^{(t)} - \theta^*\|
\end{equation}
where $\theta^*$ is the optimal policy. By improving $\kappa$, TopoCurate accelerates RL training.

\paragraph{Sample Efficiency Bounds.}
For standard task selection with uniform sampling, the sample complexity to achieve $\epsilon$-optimal policy is:
\begin{equation}
N_{\text{tasks}} = O\left(\frac{|\mathcal{A}|^2 |\mathcal{S}|^2}{\epsilon^2 (1-\gamma)^4}\right)
\end{equation}
where $|\mathcal{A}|, |\mathcal{S}|$ are the action and state space sizes, and $\gamma$ is the discount factor. TopoCurate reduces this by a factor proportional to the average SNR:
\begin{equation}
N_{\text{tasks}}^{\text{TopoCurate}} = O\left(\frac{|\mathcal{A}|^2 |\mathcal{S}|^2}{\epsilon^2 (1-\gamma)^4 \cdot \mathbb{E}[\text{SNR}]}\right)
\end{equation}

\paragraph{Robustness to Task Distribution Shift.}
In deployment, the test task distribution $p_{\text{test}}(\mathcal{T})$ may differ from training. Tasks with high structural diversity (high $V_{\text{div}}$) provide better generalization. Formally, the generalization gap is bounded by:
\begin{equation}
\begin{aligned}
\mathbb{E}_{\mathcal{T}\sim p_{\mathrm{test}}}[J(\theta)]
&-
\mathbb{E}_{\mathcal{T}\sim p_{\mathrm{train}}}[J(\theta)] \\
&\le C\cdot D_{\mathrm{KL}}(p_{\mathrm{test}}\parallel p_{\mathrm{train}})
\end{aligned}
\end{equation}
TopoCurate minimizes this gap by training on structurally diverse tasks, reducing the effective $D_{\text{KL}}$ through improved coverage of the task manifold.

\paragraph{Robustness in Low Data Regimes:} In scenarios where trajectory data is sparse, the topological structure may be underdeveloped. In such cases, TopoCurate naturally degrades to a uniform sampling strategy (as $\Phi(v)$ becomes uniform), providing a \textit{safe fallback mechanism}.

\newpage 

\section{Case Studies}
\subsection{Trajectory Comparison for Agent Behaviours }

At the trajectory level, we compare multiple successful trajectories within the same task.
Regarding diversity, different correct trajectories for the same task exhibit distinct step-by-step procedures and reasoning paths, such as what information is checked first, how the problem is decomposed, and how the agent transitions between subtasks. 
\begin{table*}[htbp]  
\begin{trainingexample}{Trajectory Comparison \,|\, \texttt{Diversity}}
\renewcommand{\arraystretch}{1.4}
\begin{tabularx}{\linewidth}{>{\raggedright\arraybackslash}p{2.3cm} X}
  \EXrow{ExpertBG}{Task:}{
    Change the existing flight/reservation from JFK to LAX to a flight to the Bay Area (SFO or OAK), and strongly prefers not to cancel and rebook.
    }
  \EXrow{StateBG}{Trajectory 1:}{
    Treat the issue as a "rules/system boundaries" problem. First, clearly explain whether the destination can be changed, and then move on to the next steps.
    }
  \EXrow{lightgray}{Typical steps:}{
    Collect identity and order information → Confirm facts/background → Clarify hard constraints → Provide actionable solutions within the constraints
    }
  \EXrow{StateBG}{Trajectory 2: }{
    Quickly reject infeasible requests, then immediately provide the information needed for the user's decision (flights/prices).
    }
  \EXrow{lightgray}{Typical steps:}{
    Collect identity and order information → Provide a quick conclusion → Immediately present alternative solutions → Give guidance for the next step
    }
  \EXrow{SRBG}{Comparison:}{
Trajectory 1: First, explain the rules thoroughly, then proceed.

Trajectory 2: Directly give the conclusion, then provide the alternative flight prices.
    }
\end{tabularx}
\end{trainingexample}
\end{table*}
Regarding efficiency, we contrast pairs of trajectories that share the same starting conditions and reach the same successful outcome. Despite achieving identical results, they differ in their intermediate steps and overall length, reflecting distinct ways of organizing the process like using a single consolidated processing loop versus a phased multi-turn progression.

\subsection{Task Comparison for Training Potential Evaluation}
At the task level, we observe that even when two tasks have a similar number of correct trajectories or comparable success rates, they may still show significant differences in the number of valid Solution Families they contain. This reveals the varying degrees of convergence or diversity manifested in the overall structure of the tasks.

\begin{table*}[htbp]
\begin{trainingexample}{Trajectory Comparison \,|\, \texttt{Efficiency}}
\renewcommand{\arraystretch}{1.4}
\begin{tabularx}{\linewidth}{>{\raggedright\arraybackslash}p{2.3cm} X}
  \EXrow{ExpertBG}{Task:}{
The user needs to modify the pending order, changing the T-shirt size from M to L and the shoe color from black to blue. Confirm the price difference and use the \$75 gift card balance to cover any extra cost.
    }
  \EXrow{IWMBG}{Current State:}{
The user wants to modify the pending order and has completed email verification. The next step is to locate the order and process the changes and price difference payment.
    }
  \EXrow{StateBG}{Trajectory 1:}{
After email verification, merge the process for a single cycle, shorter.
    }
  \EXrow{lightgray}{Typical steps:}{
\textcolor{blue}{Email verification} → Locate pending order → Check product variants / calculate price difference → Confirm gift card balance can cover → User confirmation → \textcolor{blue}{Submit modification and return success result}
    }
  \EXrow{StateBG}{Trajectory 2: }{
After email verification, proceed in phases, with additional calculation steps, resulting in multiple cycles, longer.
    }
  \EXrow{lightgray}{Typical steps:}{
\textcolor{blue}{Email verification} → Output order status / summary and confirm pending → User clarifies modification request → Check product variants + additional price difference calculation → Confirm gift card availability → User confirmation → \textcolor{blue}{Submit modification and return success result}
    }
  \EXrow{SRBG}{Comparison:}{
With the same user introduction and goal,
  
Trajectory 1: Combine processing (locating the order and addressing modification requests in one step), making it shorter.

Trajectory 2: Use "phased processing (first confirming the order, then having the user restate their request) + additional calculation steps," making it longer.
    }
\end{tabularx}
\end{trainingexample}
\end{table*}

\begin{table*}[htbp]
\begin{trainingexample}{Task Comparison}
\renewcommand{\arraystretch}{1.4}
\begin{tabularx}{\linewidth}{>{\raggedright\arraybackslash}p{2.3cm} X}
  \EXrow{ExpertBG}{Task 1:}{
    The user is unable to send MMS (multimedia messages) and needs to troubleshoot and restore functionality.
    }
  \EXrow{StateBG}{Solution type:}{
Type 1: Connectivity → permissions closure: restore connectivity first (enable mobile data, reseat SIM, etc.), then identify missing SMS permission and enable it to fix MMS.

Type 2: SIM-first → permissions convergence: fix SIM/no-signal issues first, then enable missing SMS permission to resolve MMS.
    }
  \EXrow{ExpertBG}{Task 2:}{
    The user is in France and experiencing slow or intermittent mobile data, requesting "excellent speed."
    }
  \EXrow{StateBG}{Solution type: }{
Type 1: Usage/limit-first + settings optimization: refuel/add data first, then enable data roaming and disable Data Saver, reaching “excellent” speed in a speed test.

Type 2: Settings-first (roaming → data saver): enable Data Roaming first, then disable Data Saver to improve speed.

Type 3: Settings-first (data saver → roaming): disable Data Saver first, then enable Data Roaming.

Type 4: Recovery/backtracking type: after a setting change causes loss of connectivity or user anxiety, proceed with step-by-step checks/restart, and still end up with the combo fix.
    }
  \EXrow{SRBG}{Comparison:}{
Task 1: 8 correct trajectories, 2 types, more "convergent" (most correct trajectories fall into a few ending shapes).

Task 2: 8 correct trajectories, 4 types, ending shapes are more dispersed (more types).
    }
\end{tabularx}
\end{trainingexample}
\end{table*}

\section{E Ethics and Resource Statement}
\paragraph{Dataset.}
Our experiments use public benchmarks including Tau2Bench, BFCLv3 and BFCLv4.
Synthetic task trajectories are adapted from the APIGen‑MT pipeline on the Tau2 environment. No human annotations are collected.

\paragraph{Models.}
We use Qwen3‑8B / Qwen3‑32B‑Thinking as our backbone model.
Claude‑4.5‑Sonnet is used for distilling seed expert trajectories.
Embedding models include Jina‑embeddings‑v2‑base‑en and Qwen3‑0.6B‑Embedding.

\paragraph{LLM Usage.}
Large‑language models are used for paper writing polishing only and not for generating core experimental results.

\paragraph{Computation.}
Experiments are conducted on GPU clusters for SFT and GRPO reinforcement‑learning training.


\end{document}